\documentclass[lettersize,journal]{IEEEtran}

\usepackage{amsmath,amsfonts}
\usepackage{algorithmic}
\usepackage{array}
\usepackage[caption=false,font=normalsize,labelfont=sf,textfont=sf]{subfig}
\usepackage{textcomp}
\usepackage{stfloats}
\usepackage{url}
\usepackage{verbatim}
\usepackage{graphicx}
\usepackage{epstopdf}

\def\BibTeX{{\rm B\kern-.05em{\sc i\kern-.025em b}\kern-.08em
    T\kern-.1667em\lower.7ex\hbox{E}\kern-.125emX}}
\usepackage{balance}
\usepackage{siunitx}

\usepackage{amsmath}
\usepackage{amssymb}
\usepackage{booktabs} 
\usepackage{multirow}
\usepackage{pifont} 

\usepackage[table]{xcolor} 
\usepackage{ragged2e} 
\usepackage{tabularx}

\usepackage[percent]{overpic} 

\newcolumntype{L}[1]{>{\raggedright\arraybackslash}p{#1}}

\usepackage{comment}

\begin{document}

\title{DAS-SK: An Adaptive Model Integrating Dual Atrous Separable and Selective Kernel CNN for Agriculture Semantic Segmentation}
\author{Mei Ling Chee, Thangarajah Akilan \IEEEmembership{SMIEEE}, 
Aparna Ravindra Phalke \IEEEmembership{IEEE}, Kanchan Keisham
\thanks{ML. Chee is a graduate with the Electrical \& Computer Engineering Department of Lakehead University, Thunder Bay, ON, Canada.}
\thanks{T. Akilan is with the Faculty of Software Engineering of Lakehead University, Thunder Bay, ON, Canada.}
\thanks{AR. Phalke is with the Faculty of Applied Science of the University of Alabama, Huntsville, AL, USA.}
\thanks{K. Keisham is with the School of Computer Science Engineering and Information Systems, Vellore Institute of Technology, Tamil Nadu, India.}
}

\markboth{DAS-SK: An Adaptive Model Integrating Dual
Atrous Separable and Selective Kernel CNN for
Agriculture Semantic Segmentation (PREPRINT)}%
{Shell \MakeLowercase{\textit{et al.}}: A Sample Article Using IEEEtran.cls for IEEE Journals}
\maketitle

\begin{abstract}
Semantic segmentation in high-resolution agricultural imagery demands models that strike a careful balance between accuracy and computational efficiency to enable deployment in practical systems. In this work, we propose DAS-SK, a novel lightweight architecture that retrofits selective kernel convolution (SK-Conv) into the dual atrous separable convolution (DAS-Conv) module to strengthen multi-scale feature learning. The model further enhances the atrous spatial pyramid pooling (ASPP) module, enabling the capture of fine-grained local structures alongside global contextual information. Built upon a modified DeepLabV3 framework with two complementary backbones—MobileNetV3-Large and EfficientNet-B3, the DAS-SK model mitigates limitations associated with large dataset requirements, limited spectral generalization, and the high computational cost that typically restricts deployment on UAVs and other edge devices. Comprehensive experiments across three benchmarks: LandCover.ai, VDD, and PhenoBench, demonstrate that DAS-SK consistently achieves state-of-the-art performance, while being more efficient than CNN-, transformer-, and hybrid-based competitors. Notably, DAS-SK requires up to 21× fewer parameters and 19× fewer GFLOPs than top-performing transformer models. These findings establish DAS-SK as a robust, efficient, and scalable solution for real-time agricultural robotics and high-resolution remote sensing, with strong potential for broader deployment in other vision domains. Code is available at https://github.com/irene7c/DAS-SK.git.
\end{abstract}


\section{Introduction}

\IEEEPARstart{T}{he} rapid development of remote sensing technologies has transformed smart agriculture by enabling large-scale and precise monitoring of farmlands. High-resolution imagery from satellites and unmanned aerial vehicles (UAVs) now provides unprecedented opportunities for assessing crop growth, monitoring field health, and supporting precision farming practices. Extracting actionable information from such imagery, however, remains challenging due to the need for accurate interpretation of complex spatial and spectral patterns.

Semantic segmentation has emerged as a powerful solution by providing pixel-level classification that generates fine-grained maps of crop and land cover distributions~\cite{ma2024multilevel, ma2025unified, li2025lightweight, he2022swin, suresh2025ss}. These outputs are critical for applications such as land type identification, weed detection, crop classification, and yield estimation. Nevertheless, agricultural images differ significantly from urban or natural scenes, with challenges including high intra-class variability from crop phenology and management practices, seasonal variability, illumination changes, occlusions, and the heavy computational demands of very high-resolution data~\cite{tinysegformer2024,wu2025semantic}.
Traditional thresholding and machine learning approaches offered early solutions but were heavily dependent on handcrafted features, limiting their adaptability to diverse conditions~\cite{campos2016multitemporal}. The introduction of deep learning, particularly convolutional neural networks (CNNs), brought major improvements. Architectures such as U-Net~\cite{ronneberger2015u}, DeepLab~\cite{chen2017deeplab}, and SegNet~\cite{badrinarayanan2017segnet} became widely used, demonstrating strong capabilities in capturing spatial structures and delineating object boundaries in agricultural scenes.

More recently, transformer-based models~\cite{dosovitskiy2020image,zheng2021rethinking} have attracted attention for their ability to model long-range dependencies and global context. While these models outperform CNNs in contextual reasoning, their computational complexity makes real-time deployment on UAVs and edge devices impractical~\cite{ma2024multilevel}. Hybrid CNN–Transformer architectures have been proposed to balance local and global feature modeling, but finding an optimal trade-off between accuracy, efficiency, and scalability remains an open challenge~\cite{abid2025contextformer}.

Many existing models perform well in controlled conditions but generalize poorly across diverse agricultural scenes. Furthermore, the high computational cost of state-of-the-art architectures restricts their deployment in precision agriculture systems, which require lightweight, energy-efficient, and real-time solutions~\cite{niu2022hsi}. These challenges highlight the need for innovative segmentation models that strike a balance between high accuracy and improved efficiency and adaptability.
In this direction, this work delivers a significant contribution by harnessing the complementary strengths of DAS-Conv and SK-Conv within a dual-stream architecture built on MobileNetV3-Large~\cite{chen2017rethinking} and EfficientNet-B3~\cite{tan2019efficientnet}. This design not only boosts segmentation accuracy but also achieves remarkable computational efficiency, paving the way for scalable and real-time semantic segmentation on resource-constrained edge devices.
Thus, the main contributions of DAS-SK can be summarized as follows:
\begin{enumerate}
    \item A lightweight segmentation architecture with multi-scale feature representation and adaptive receptive fields for robust crop/non-crop discrimination.
    \item Introduction of an efficient context aggregation mechanism that captures both fine local details and global structure without heavy computation.
    \item Evaluation on multiple benchmark datasets to demonstrate superior balance between accuracy and efficiency of the proposed model compared to existing approaches. 
\end{enumerate}

The organization of the rest of the paper is as follows: Section~\ref{Sec-related-works} reviews the relevant literature, providing context for this work. Section~\ref{Sec-methodology} outlines the proposed approach and its novelty. Section~\ref{Sec-experiments} describes the experimental setup, presents the results, and analyzes the model's performance. Section~\ref{Sec-Conclusion} summarizes the findings and offers ideas for future research.

\section{Related Works} \label{Sec-related-works}
\noindent

\begin{table*}[!ht]
\caption{A comparison of key semantic segmentation related works in the agricultural field}\label{table-literature-review} \vspace{-0.2cm}
\begin{center}
\setlength{\tabcolsep}{1.0pt}
\renewcommand{\arraystretch}{1}
\begin{tabular}{p{0.2cm} |L{3.4cm} |L{0.25\linewidth} | L{0.24\linewidth} | L{0.28\linewidth} }
\toprule
& \multicolumn{1}{c}{\textbf{Model}} 
& \multicolumn{1}{c}{\textbf{Application}} 
& \multicolumn{1}{c}{\textbf{Approach}} 
& \multicolumn{1}{c}{\textbf{Limitations}} \\
\hline\hline
\multirow{3}{*}{\rotatebox{90}{\hspace{-0.7cm}CNN}}  & U-Net MobileVit-S~\cite{HERNANDEZ202515} & Downy mildew detection in grapevines & Transfer learning, data augmentation & Small dataset; lighting/symptom variability\\
& Residual U-Net (ResNet101)~\cite{garibaldi2025leveraging} & Corn crop, weed segmentation & Transfer learning, data augmentation & Large datasets needed; narrow-leaf weeds confused with soil; high computation \\ 
& DeepLabV3 MobileNetV3~\cite{ling2025dual} & Farmland anomaly detection & Dual atrous separable convolutions, optimized dilation, skip connections & Limited experimental study on a single dataset \\ 
\midrule
\multirow{3}{*}{\rotatebox{90}{\hspace{-1cm} Transformer}} 
 & SegFormer (MiT-B0/B3/B5)~\cite{Elmessery2024Semantic} & Strawberry disease segmentation & Data augmentation, SAM-assisted annotation & Limited dataset diversity; environment-sensitive \\ 
 & MiT-B3~\cite{shen2022aaformer} & Farmland anomaly detection & Boundary map fusion in encoder & Minor improvement; needs boundary maps \\ 
& SegFormer + MiT-B5~\cite{tavera2022augmentation}  & Agricultural aerial image segmentation & Adaptive sampling, augmentation invariance & Small-class pixel imbalance reduces efficiency \\ 
\midrule
\multirow{3}{*}{\rotatebox{90}{\hspace{-0.5cm}Hybrid}} 
 & TinySegformer~\cite{tinysegformer2024} & Pest detection on edge devices & Sparse attention, pruning & Dataset diversity; device resource limits \\ 
 & FireViTNet (MobileViT)~\cite{wang2024firevitnet} & Wildland fire detection & CBAM, Dense ASPP, spatial pooling & GPU/memory constraints; fixed resolution \\ 
 & CNN–ViT hybrid~\cite{wei2025hybrid} & Crop and weed detection on embedded devices & Encoder-decoder fusion, global-local semantic fusion & Balancing accuracy and efficiency in complex fields \\ 
\hline\hline
\end{tabular} \vspace{-0.2cm}
\end{center}
\end{table*}

The key studies in agricultural semantic segmentation can be grouped into three: CNN-based, transformer-based, and hybrid models. Table~\ref{table-literature-review} summarizes the main details of these studies.

\begin{figure*}[!ht]
\centering   
\includegraphics[trim={0.00cm, 23.8cm, 0.00cm, 0.0cm}, clip, width=1.0\textwidth]{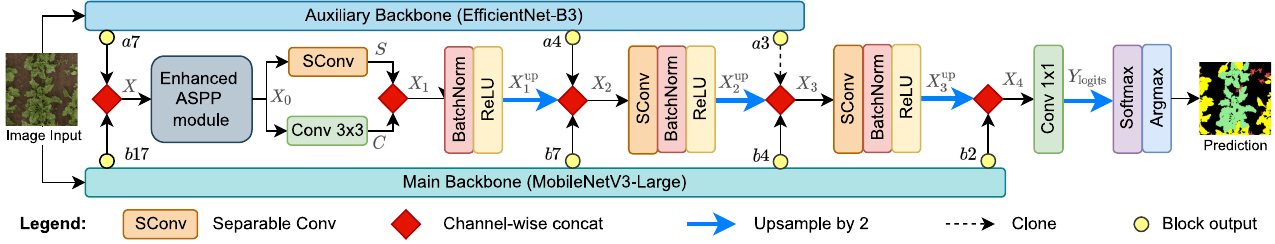} \vspace{-0.4cm}
\caption{The overall architecture. The backbones extract multi-scale features from the input, which are then fused and refined via the enhanced ASPP with \texttt{DAS-SKConv} (cf.~Fig.~\ref{fig-DAS-SKConv}). The decoder then progressively upsamples and refines the fused features with skip connections to produce accurate segmentation.}
\label{fig-overall-architecture} \vspace{-0.3cm}
\end{figure*}

\textbf{CNN-based Approaches:} 
Semantic segmentation in agricultural and remote sensing tasks demands models balancing accuracy and computational efficiency. CNNs remain practical for UAV and edge deployment due to their lightweight nature and real-time capability~\cite{10843732,11121589,10440354}. Hernández~\textit{et~al.}~\cite{HERNANDEZ202515} used U-Net with a MobileViT-S encoder for early vineyard disease detection, achieving strong performance via transfer learning and tailored augmentations. Garibaldi-Márquez~\textit{et~al.}~\cite{garibaldi2025leveraging} applied a residual U-Net with ResNet backbones for corn–weed segmentation, outperforming Mask R-CNN but struggling with thin weed structures in dense canopies. By integrating optimized skip connections and atrous separable convolutions within a DeepLabV3-based framework, the model improves efficiency and generalization, though it still struggles with anomaly classes in complex datasets~\cite{ling2025dual}. Collectively, these studies confirm that CNNs remain the backbone of agricultural semantic segmentation, while trade-offs between efficiency and accuracy persist.

\textbf{Transformer-based Approaches:} Transformers have advanced agricultural vision tasks by capturing long-range dependencies and global context. Elmessery~\textit{et~al.}~\cite{Elmessery2024Semantic} showed that MiT-B3/B5 outperform MiT-B0, though performance remains dataset- and condition-sensitive. Shen~\textit{et~al.}~\cite{shen2022aaformer} introduced AAFormer, combining MiT and SE modules with boundary-aware decoding, improving feature extraction but increasing input complexity. Tavera~\textit{et~al.}~\cite{tavera2022augmentation} tackled class imbalance using adaptive sampling, which enhanced segmentation for major classes but reduced accuracy for rare ones. Despite the strengths, transformers are computationally demanding and data hungry, limiting practical deployment.

\textbf{Hybrid Approaches:} 
Some models merge CNNs’ efficiency with transformers’ contextual reasoning~\cite{akilan2025self}. For instance, Zhang \textit{et~al.}~\cite{tinysegformer2024} proposed TinySegFormer, a sparse-attention framework enabling real-time pest detection. 
Wang~\textit{et~al.}~\cite{wang2024firevitnet} combined MobileViT, CBAM, Dense ASPP, and SP pooling for forest fire detection, achieving strong accuracy but with high memory demands. 
Meanwhile, Wei~\textit{et~al.}~\cite{wei2025hybrid} developed a lightweight hybrid network for weeding robots, reaching high accuracy, illustrating the potential of efficient design.

In short, CNNs remain the practical choice for real-time systems, with low computational cost and adaptability to UAV-based platforms. Transformers show superior global reasoning but require substantial resources, while hybrid models balance both with moderate complexity. Building on these insights, the proposed DAS-SK enhances multi-scale feature representation and adaptability, addressing the persistent trade-off between accuracy and efficiency in agricultural semantic segmentation.

\section{Methodology}
\label{Sec-methodology}
\noindent
Fig.~\ref{fig-overall-architecture} depicts the proposed model adopting a dual-backbone encoder–decoder architecture composed of 3 key components: dual backbone, an enhanced ASPP module, and a hierarchical decoder network. Their detailed mathematical expressions are presented in the Supplementary Materials. 

\subsection{The Dual Backbone}
\noindent
MobileNetV3-Large (cf.~Table~\ref{table-main-backbone}) is used as the main backbone, which consists of a series of convolution (Conv) and inverted residual blocks that progressively reduce the spatial resolution of the input while increasing the number of learned feature maps. 
The backbone's early layers focus on low-level details, such as edges and textures, while the deeper layers encode high-level semantic concepts and object boundaries.   
The auxiliary backbone is a truncated version of EfficientNet-B3, used to extract complementary cues. 
It is devised by taking the first six blocks of EfficientNet-B3’s layers and then adding a Conv layer to adjust the output channels (cf.~Table~\ref{table-aux-backbone}). 

\begin{table}[!tp]
\caption{Details of the primary backbone (Mobilenetv3-large)}
\label{table-main-backbone} \vspace{-0.2cm}
\begin{center}
\setlength{\tabcolsep}{1.5pt} 
\begin{tabular}{ccc}
\toprule 
Input $(\mathbin{H\!\times\!} W \!\mathbin{\!\times\!} \!C)$ & $f(k, s, a)$ & Output $(\mathbin{H\!\times\!} W \!\mathbin{\!\times\!} \!C)$ \\
\midrule
$512\!\times\!512\!\times\!3$ & \texttt{Conv(3,2,HardSwish)} & $256\!\times\! 256\!\times\!16$ \\
~$256\!\times\!256\!\times\!16$ & \texttt{Bneck(3,1,ReLU)} & ~~~~~~$256\!\times\! 256\!\times\!16$ \textcolor{teal}{[b2]} \\
~$256\!\times\!256\!\times\!16$ & \texttt{Bneck(3,2,ReLU)} & $128\!\times\!128\!\times\!24$ \\
$128\!\times\!128\!\times\!24$ & \texttt{Bneck(3,1,ReLU)} & ~~~~~~$128\!\times\!128\!\times\!24$ \textcolor{teal}{[b4]} \\
$128\!\times\!128\!\times\!24$ & \texttt{Bneck(5,2,ReLU)} & $64\!\times\!64\!\times\!40$ \\
$64\!\times\!64\!\times\!40$ & \texttt{Bneck(5,1,ReLU)} & $64\!\times\!64\!\times\!40$ \\
$64\!\times\!64\!\times\!40$ & \texttt{Bneck(5,1,ReLU)} & ~~~~~~$64\!\times\!64\!\times\!40$ \textcolor{teal}{[b7]} \\
$64\!\times\!64\!\times\!40$ & \texttt{Bneck(3,2,HardSwish)} & $32\!\times\!32\!\times\!80$ \\
$32\!\times\!32\!\times\!80$ & \texttt{Bneck(3,1,HardSwish)} & $32\!\times\!32\!\times\!80$ \\
$32\!\times\!32\!\times\!80$ & \texttt{Bneck(3,1,HardSwish)} & $32\!\times\!32\!\times\!80$ \\
$32\!\times\!32\!\times\!80$ & \texttt{Bneck(3,1,HardSwish)} & $32\!\times\!32\!\times\!80$ \\
$32\!\times\!32\!\times\!80$ & \texttt{Bneck(3,1,HardSwish)} & $32\!\times\!32\!\times\!112$ \\
$32\!\times\!32\!\times\!112$ & \texttt{Bneck(3,1,HardSwish)} & $32\!\times\!32\!\times\!112$ \\
$32\!\times\!32\!\times\!112$ & \texttt{Bneck(5,2,HardSwish)} & $32\!\times\!32\!\times\!160$ \\
$32\!\times\!32\!\times\!160$ & \texttt{Bneck(5,1,HardSwish)} & $32\!\times\!32\!\times\!160$ \\
$32\!\times\!32\!\times\!160$ & \texttt{Bneck(5,1,HardSwish)} & $32\!\times\!32\!\times\!160$ \\
$32\!\times\!32\!\times\!160$ & \texttt{Conv(1,1,HardSwish)} & ~~~~~~~$32\!\times\!32\!\times\!480$ \textcolor{teal}{[b17]} \\
\midrule
\end{tabular}
\begin{tabularx}{\linewidth}{@{}>{\raggedright\arraybackslash}X@{}}
Total \# of trainable parameters: 2.894 M; $f(k, s, a)$, where $f(\cdot)$ - block type, $k$ - kernel size, $s$ - stride, $a$ - activation function, and Bneck - bottleneck block, annotations in the output column refer to the residual block output utilized in the decoder, as illustrated in Fig.~\ref{fig-overall-architecture}.\\ 
\bottomrule
\end{tabularx} \vspace{-0.2cm}
\end{center}
\end{table}

\subsection{The DAS-SKConv Module}

\begin{figure*}[!t]
\centering  
\includegraphics[trim={0.00cm, 21.60cm, 0.00cm, 0.0cm}, clip, width=1.0\textwidth]{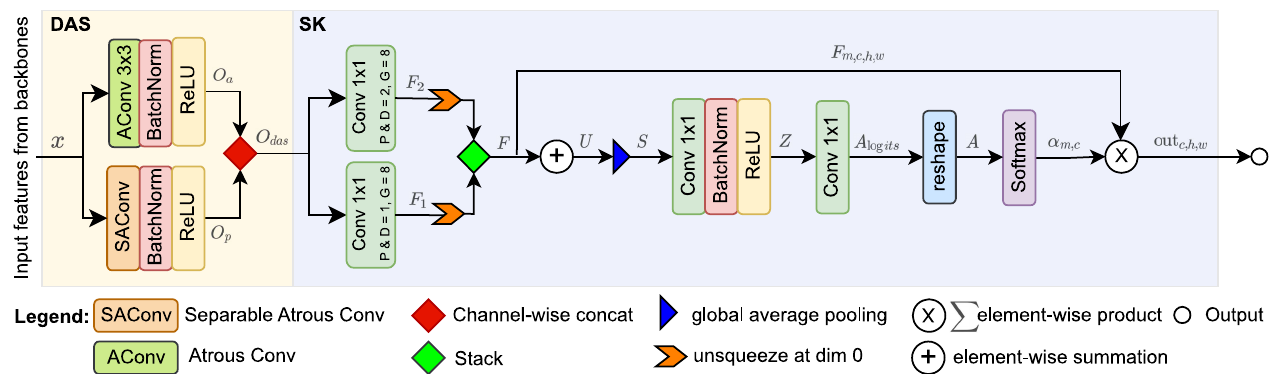} \vspace{-0.4cm}
\caption{The proposed \texttt{DAS-SKConv} module. The DAS block combines atrous separable and standard atrous convolutions to capture both fine and broad spatial features. The SK attention mechanism adaptively weights multi-branch features through channel-wise attention, producing context-aware representations.}
\label{fig-DAS-SKConv} \vspace{-0.2cm}
\end{figure*}

\noindent
It is designed to enhance feature extraction by combining DAS convolutions with SK attention, as shown in Fig.~\ref{fig-DAS-SKConv}. The DAS block employs two parallel convolutional paths: a separable atrous branch for efficient channel mixing and a standard atrous branch for spatial context expansion. Their outputs are concatenated to form rich multi-scale features, which are then refined through SK attention. This attention mechanism adaptively reweights channels across multiple receptive fields, enabling dynamic emphasis on the most informative spatial scales. By integrating multiscale spatial sampling with adaptive channel attention, \texttt{DAS-SKConv} produces context-aware representations that improve both local detail preservation and global consistency in semantic segmentation tasks.

\subsection {Enhanced ASPP module}
\label{subsection-assp}

\begin{figure}[!tp]
\centering
\begin{overpic}[trim={0.5cm 13.7cm 0.0cm 0.0cm}, clip, width=0.95\columnwidth]{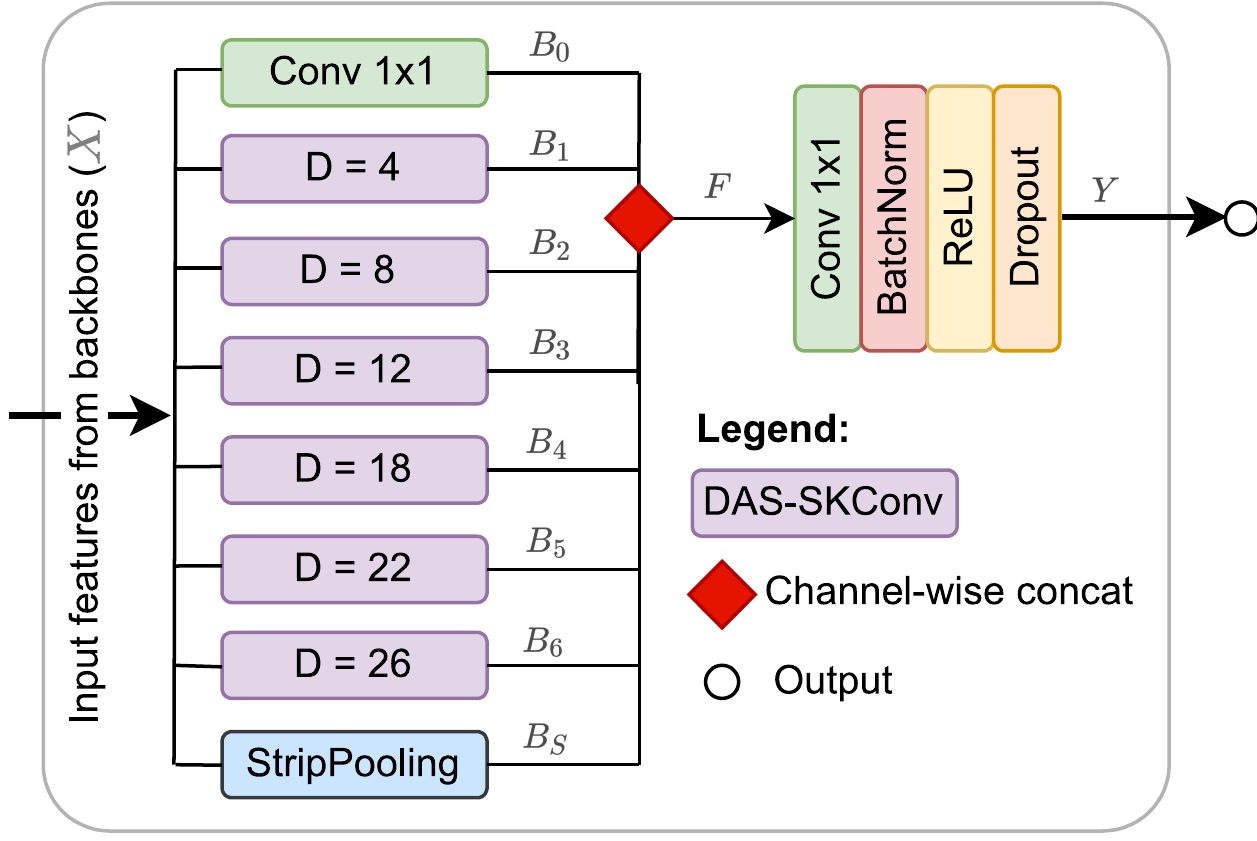}
  \put(55,4){\scriptsize \shortstack[r]{
$D$ - dilation rates of the \\ DAS-SKConv
in Fig.~\ref{fig-DAS-SKConv}.}}
\end{overpic} \vspace{-0.2cm}
\caption{The enhanced ASPP module. High-dimensional backbone features are processed via parallel branches, including a $1\times1$ Conv, six \texttt{DAS-SKConv} with varying dilation rates, and a strip pooling branch.
}
\label{fig-enhanced-ASPP-module} \vspace{-0.2cm}
\end{figure}

Fig.~\ref{fig-enhanced-ASPP-module} depicts the enhanced ASPP. It enriches feature representations by capturing multi-scale contextual information from high-dimensional backbone features. It employs parallel branches consisting of a $1\times1$ Conv for channel reduction, six \texttt{DAS-SKConv} layers with varying dilation rates $\{4,8,12,18,22,26\}$ for multi-scale receptive fields, and a strip pooling branch to model long-range dependencies along horizontal and vertical directions—effective for structured agricultural patterns. The outputs are concatenated and compressed via a $1\times1$ Conv followed by batch normalization, ReLU, and dropout. This design effectively integrates local detail, global structure, and adaptive scale awareness, producing compact yet context-rich features optimized for semantic segmentation.

\subsection{Decoder}

The decoder, as shown in Fig.~\ref{fig-overall-architecture}, fuses high-level features ($a7$ and $b17$) from both backbones into $X \in \mathbb{R}^{960\times H\times W}$, which is refined through the Enhanced ASPP module to produce $X_0 \in \mathbb{R}^{256\times H\times W}$. Two parallel branches—a separable Conv and a standard $3\times3$ Conv—extract complementary spatial information, and their outputs are concatenated, normalized, and activated. The refined features are progressively upsampled and fused with corresponding intermediate backbone outputs ($a4$, $b7$, $a3$, $b4$, $b2$) to recover spatial detail through hierarchical skip connections. Each fusion stage is followed by separable Conv, batch normalization, and ReLU activation for feature refinement. Finally, a $1\times1$ Conv projects the representation into $C$ class channels, and an upsampling operation restores full spatial resolution. Applying Softmax and argmax yields the final pixel-level segmentation map.

\begin{table}[!tp]
\setlength{\tabcolsep}{3pt} 
\caption{Details of the auxiliary backbone (EfficientNet-B3)}
\label{table-aux-backbone} \vspace{-0.2cm}
\begin{center}
\setlength{\tabcolsep}{6pt} 
\begin{tabular}{ccc}
\toprule 
Input $(\mathbin{H\!\times\!} W \!\mathbin{\!\times\!} \!C)$& $f(k, l)$ & Output $(\mathbin{H\!\times\!} W \!\mathbin{\!\times\!} \!C)$\\
\midrule
$512\times512\times3$ & \texttt{Conv(3,1)} & $256\times256\times40$ \\
$256\times256\times40$ & \texttt{MBConv1(3,2)} & $256\times256\times24$ \\
$256\times256\times24$ &\texttt{MBConv6(3,3)} & ~~~~~$128\times128\times32$ \textcolor{teal}{[a3]} \\
$128\times128\times32$ & \texttt{MBConv6(5,3)} & ~~~~~$64\times64\times48$ \textcolor{teal}{[a4]} \\
$64\times64\times48$ &\texttt{MBConv6(3,5)} & $32\times32\times96$ \\
$32\times32\times96$ & \texttt{MBConv6(5,5)} & ~$32\times32\times136$  \\
$32\times32\times136$ & \texttt{Conv(1,1)} & ~~~~~~~$32\times32\times480$ \textcolor{teal}{[a7]} \\
\midrule
\end{tabular}
\begin{tabularx}{\linewidth}{@{}>{\raggedright\arraybackslash}X@{}}
Total \# of trainable parameters: 2.256 M; $f(k, l)$, where $f(\cdot)$ - block type, $k$ - kernel size, $l$ - Number of time this Conv appears in sequence, annotations in the output column refer to the residual block output utilized in the decoder, as illustrated in Fig.~\ref{fig-overall-architecture}.\\
\bottomrule
\end{tabularx} \vspace{-0.2cm}
\end{center}
\end{table}

\section{Experiments and Discussion}
\label{Sec-experiments}
\subsection{Datasets}

\begin{table}[!tp]
\caption{Summary of the benchmark datasets used in this study}
\label{table-datasets-summary} \vspace{-0.2cm}
\begin{center}
\setlength{\tabcolsep}{1pt} 
\begin{tabular}{lccccc}
\toprule 
\multirow{2}{*}{\centering \textbf{Dataset}} & \textbf{Res.} & \textbf{Seman.} & {\textbf{Image}} & \textbf{Total} & \textbf{Train/Validation/} \\
 &  (cm/px) & \textbf{Classes} & \textbf{Size}& \textbf{Samples} & \textbf{Test samples} \\
\midrule
LandCover.ai   & 25, 50 & 5 & $512\times512$ & 10674 & 7470/ 1602/ 1602  \\
VDD  & 1.75, 4.2 & 7 & $4000\times3000$ & 400 & 280/ 80/ 40  \\
PhenoBench  & 0.1 & 3 & $1024\times1024$ & 2872 & 1407/ 772/ 693*  \\
\bottomrule
\end{tabular}
\vspace{-.18cm}
\begin{flushleft} {\scriptsize 
* Predictions to be submitted to https://codalab.lisn.upsaclay.fr/competitions/13654. }\end{flushleft}
\end{center}
\vspace{-.5cm}
\end{table}

\noindent
As summarised in Table~\ref{table-datasets-summary}, this study uses three publicly available benchmark data sets for experimental analysis. 
The Land Cover from Aerial Imagery (LandCover.ai)~\cite{Boguszewski_2021_CVPR}, is a high-resolution aerial imagery dataset annotated with buildings, woodlands, water, and roads. 
The Varied Drone Dataset (VDD)~\cite{CAI2025104429} is a diverse UAV dataset covering multiple environments and conditions with seven semantic classes, and the PhenoBench~\cite{weyler2024phenobench} focuses on crop–weed discrimination and plant phenotyping under realistic field conditions. 

\subsection{Evaluation Metrics and Training Strategy}
\noindent
The model performances are evaluated on model efficiency. While mean Intersection over Union (mIoU) measures segmentation accuracy, efficiency provides a unified assessment that balances predictive performance with computational complexity.

mIoU is defined as:
\begin{equation}
mIoU = \frac{1}{k} \sum_{i=1}^{k} \frac{P_{ii}}{P_{i\cdot} + P_{\cdot i} - P_{ii}},
\end{equation}
where $k$ is the number of classes, $P_{ii}$ is the number of correctly predicted pixels for class $i$, and $P_{i\cdot}$ and $P_{\cdot i}$ denote the total true and predicted pixels for class $i$, respectively.

Hence, the model efficiency is defined as: 
\begin{equation} 
    \text{Efficiency} = \frac{\text{Diff}_{mIoU}}{\log(\text{Params}) \cdot \text{GFLOPs}} \times 100\%, \label{eq-efficiency} 
\end{equation} 
where $\text{Diff}_{mIoU}$, Params, and GFLOPs represent the mIoU difference from the baseline, the number of parameters, and the computational cost, respectively. 
Logarithmic scaling of parameter count normalizes exponential growth in model complexity, preventing large architectures from dominating the efficiency metric. This captures the diminishing performance gains with increasing size, yielding a fair and interpretable measure of how segmentation accuracy ($\text{Diff}_{mIoU}$) scales with computational compactness.

\begin{figure*}[!t]
    \centering
    \subfloat[LandCover.ai]{%
        \includegraphics[trim={0cm, 0cm, 0cm, 16.5cm}, clip, width=5.5cm]{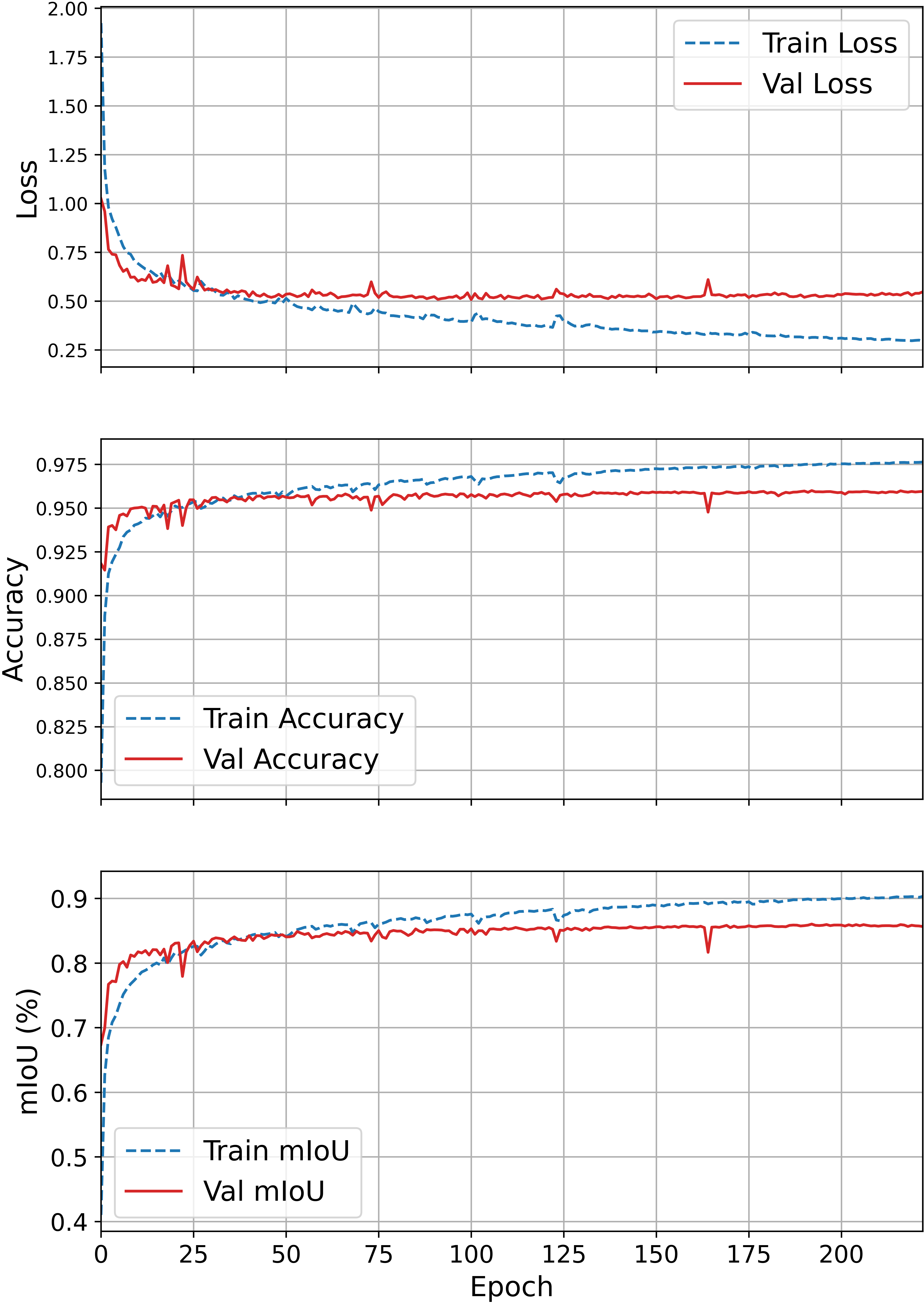}%
        \label{fig:val_curves_landcover}
    }
    \hfill
    \subfloat[VDD]{%
        \includegraphics[trim={0cm, 0cm, 0cm, 16.5cm}, clip, width=5.5cm]{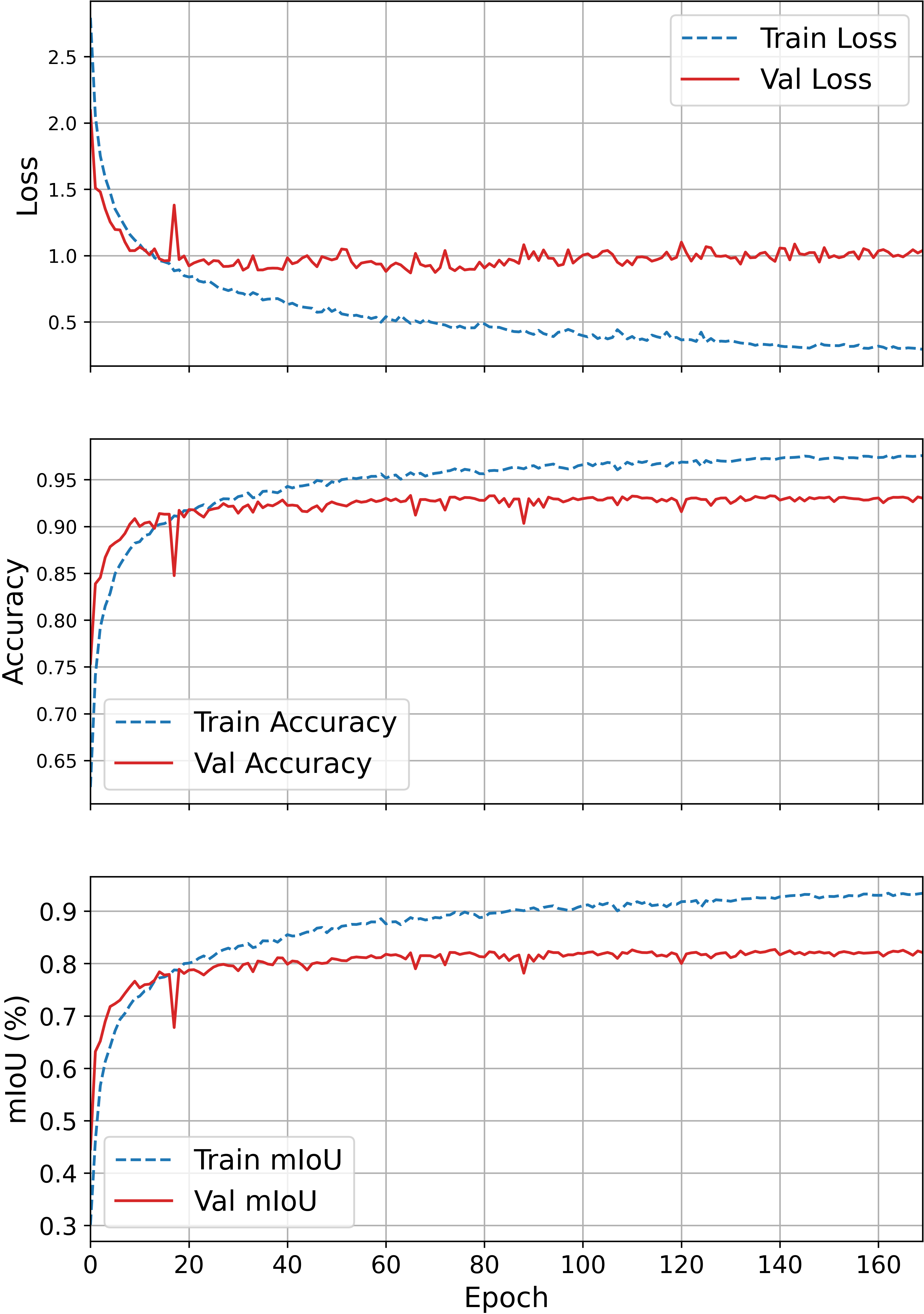}%
        \label{fig:val_curves_VDD}
    }
    \hfill
    \subfloat[PhenoBench]{%
        \includegraphics[trim={0cm, 0cm, 0cm, 16.5cm}, clip, width=5.5cm]{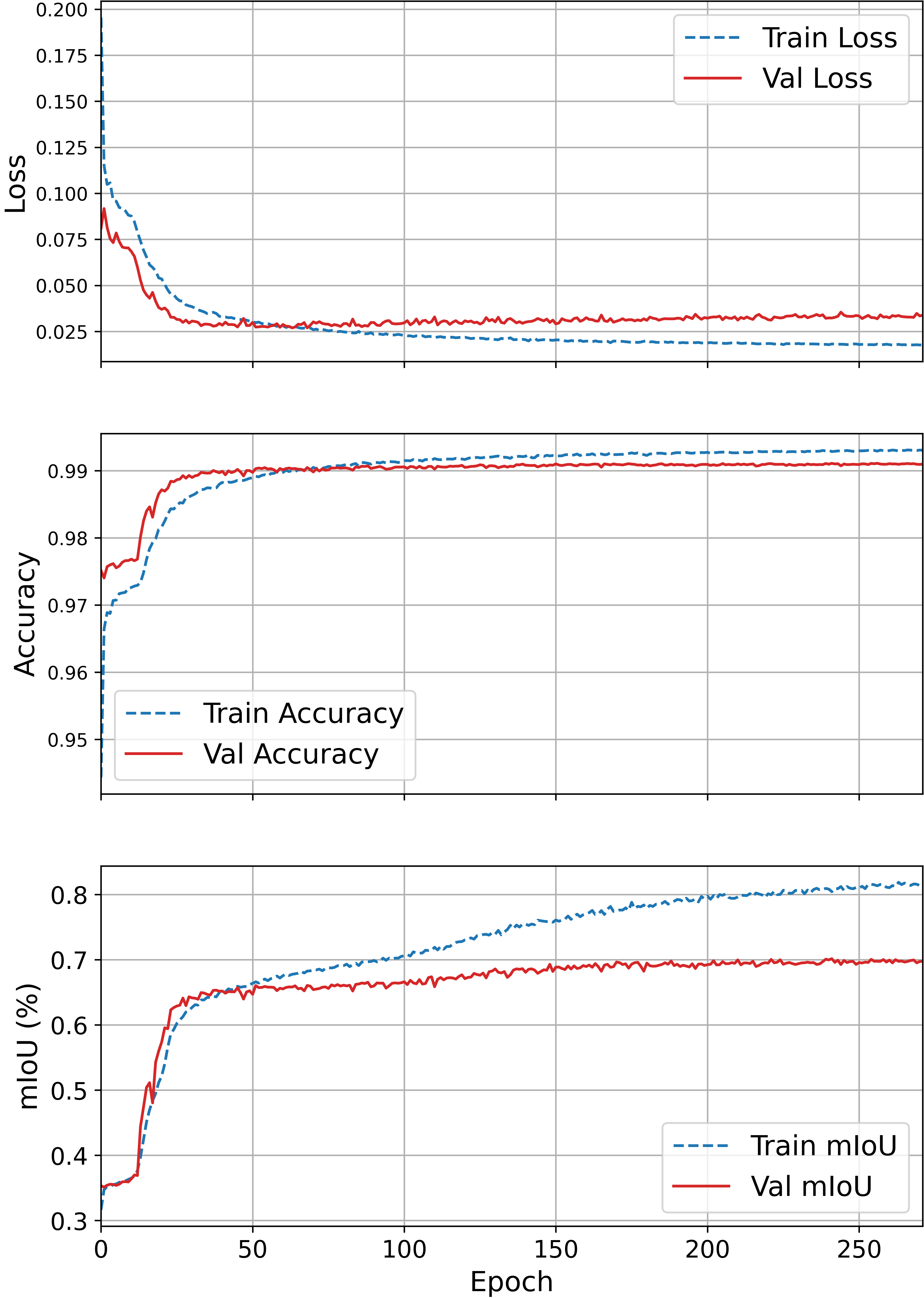}%
        \label{fig:val_curves_pheno}
    }
    \caption{Training progress of the proposed model using the configurations given in Table~\ref{table-implementation} on LandCover.ai, VDD, and PhenoBench benchmark datasets.}
    \label{fig-performance-curves}
    \vspace{-0.2cm}
\end{figure*}

Table~\ref{table-implementation} summarizes the implementation details of the proposed DAS-SK, while Fig.~\ref{fig-performance-curves} shows the model training progress on the benchmark datasets. 

\begin{table}[!tp]
\centering
\caption{Implementation details}
\label{table-implementation}
\renewcommand{\arraystretch}{1.2}
\begin{tabular}{p{1.6cm}p{6.5cm}}
\hline
\textbf{Component} & \textbf{Configuration} \\ \hline
Environment & A100-40GB GPU (Alliance Canada, Narval cluster). \\
 & NVIDIA GeForce RTX 3050 Ti GPU (to obtain FPS)\\
Framework & PyTorch 2.1 with CUDA 12.1 \\
Optimizer & SGD with learning rate = 0.001, momentum = 0.9, weight decay = 0.0005. \\
Scheduler & Cosine annealing with minimum learning rate $10^{-5}$. \\
Batch Size & 8 for input resolution of $512 \times 512$\newline 4 for larger input resolution\\
Early Stop & Patience of 30 epochs. \\
Loss Function & $\mathcal{L}_{\text{total}} = \mathcal{L}_{\text{CE}} + \mathcal{L}_{\text{Dice}} + \mathcal{L}_{\text{Focal}} + \mathcal{L}_{\text{Lovasz}}$. \\
Loss Summary & Cross Entropy (CE) – pixel accuracy; Dice – region overlap; Focal – hard samples; Lovász – mIoU optimization. \\ 
Data \newline Augmentation & Random horizontal and vertical flips, Random 90-degree rotation, ShiftScaleRotate, and color jitter.\\
\hline
\end{tabular}
\end{table}

\subsection{Quantitative Analysis}

\begin{table*}[!tp]
\setlength{\tabcolsep}{10pt} 
\caption{Model efficiency on the Landcover dataset}
\label{table-Landcover-Efficiency} \vspace{-0.4cm}
\begin{center}
\setlength{\tabcolsep}{4pt} 
\begin{tabular}{llS[table-format=2.2]SS[table-format=3.3]cccS[table-format=2.2]}
\toprule 
\multicolumn{1}{c}{\textbf{Type}} & \multicolumn{1}{c}{\textbf{Model}} & \textbf{mIoU ($\%$) $\uparrow$} & \multicolumn{1}{c}{\textbf{Param. (M) $\downarrow$}} & \multicolumn{1}{c}{\textbf{GFLOPs} $\downarrow$}  & \multicolumn{1}{c}{\textbf{Diff$_{miou}$ $\uparrow$}} & \multicolumn{1}{c}{l\textbf{og(Param.) $\downarrow$}} & \multicolumn{1}{c}{\textbf{Efficiency $(\%)$ $\uparrow$}} & \multicolumn{1}{c}{\textbf{FPS $\uparrow$}}\\
\midrule
\rowcolor{green!15} 
CNN & Our model (DAS-SK) & \underline{86.25} & ~~\textbf{10.678} & \textbf{11.25} & \underline{5.75} & \textbf{1.028} & \textbf{49.70} & 40.05 \\
CNN & HRNet \cite{yue2024ma}& 82.80 & ~~\underline{11.458} & 20.39 & 2.30 & \underline{1.059} & \underline{10.65} & 25.46\\
CNN & UNet \cite{yue2024ma}& 83.40 & 24.436 & 31.41 & 2.90 & 1.388 & 6.65 & \underline{54.34}\\
Transformer & SegFormer MiT-B2 \cite{yue2024ma}& 84.40 & 27.461 & 43.18 & 3.90 & 1.439 & 6.28 & 23.54\\
CNN & Diff-HRNet \cite{10766581} & 84.22 & 68.867 & 86.25 & 3.72 & 1.838 & 2.35 & 16.33\\
Hybrid & Ensemble UNet \cite{dimitrovski2024u} & \textbf{88.02} & 193.539 & 141.26 & \textbf{7.52} & 2.287 & 2.33 & 5.04\\
Hybrid & TransUNet \cite{yue2024ma}& 82.90 & 91.404 & 165.14 & 2.40 & 1.961 & 0.74 & 8.16\\
CNN & DeepLabV3 ResNet101 \cite{yue2024ma}& 83.00 & 58.626 & 241.14 & 2.50 & 1.768 & 0.59 & 9.59\\
Hybrid & MA-DBFAN \cite{yue2024ma} & 85.30 & 34.595 & 853.30 & 4.80 & 1.539 & 0.37 & 0.23 \\
CNN & BiSeNet \cite{yue2024ma}& 80.50 & 11.878 & \underline{12.46} & \text{Baseline} & 1.075 & \text{Baseline} & \textbf{126.65}\\
\bottomrule
\end{tabular}
\vspace{-.1cm}
\begin{flushleft} \footnotesize{Note: The Ensemble UNet consists of UNet MaxVIT-S, UNet ConvFormer-M36, and UNet EfficientNet-B7. GFLOPs and FPS are estimated for an input image size of $512 \times 512$. $\uparrow$ – higher is better; $\downarrow$ – lower is better; Boldface indicates the best result; underline indicates the 2nd-best.}\end{flushleft}
\end{center}\vspace{-0.2cm}
\end{table*}
\begin{table*}[!tp]
\caption{Model Efficiency on the VDD dataset}
\label{table-VDD-Efficiency} \vspace{-0.4cm}
\begin{center}
\setlength{\tabcolsep}{4pt} 
\begin{tabular}{llS[table-format=2.2]SS[table-format=3.3]cccS[table-format=2.2]}
\toprule 
\multicolumn{1}{c}{\textbf{Type}} & \multicolumn{1}{c}{\textbf{Model}} & \textbf{mIoU ($\%$) $\uparrow$} & \multicolumn{1}{c}{\textbf{Param. (M) $\downarrow$}} & \multicolumn{1}{c}{\textbf{GFLOPs} $\downarrow$} & \textbf{Diff$_{miou}$ $\uparrow$} & l\textbf{og(Param.) $\downarrow$} & \multicolumn{1}{c}{\textbf{Efficiency $(\%)$ $\uparrow$}}  & \multicolumn{1}{c}{\textbf{FPS $\uparrow$}}\\
\midrule
\rowcolor{green!15} 
CNN & Our model (DAS-SK) & 79.45 & ~\underline{10.678} & \underline{43.52} & ~4.08 & \underline{1.028} & \textbf{9.12} & \underline{10.33} \\
Transformer & SegFormer MiT-B2~\cite{CAI2025104429}& \textbf{85.75} & 27.461 & 164.70 & \textbf{10.38} & 1.439 & \underline{4.38} & 2.82\\
Hybrid & Mask2Former ResNet50~\cite{CAI2025104429} & 83.21 & 45.517 & 133.70 & ~7.84 & 1.658 & 3.54 & 3.48\\
Transformer & Mask2Former SwinT~\cite{CAI2025104429} & 77.85 & 47.439 & 45.74 & ~2.48 & 1.676 & 3.23 & 3.04\\
Transformer & SegFormer MiT-B5~\cite{CAI2025104429} & 82.11 & 84.708 & 180.53 & ~6.74 & 1.928 & 1.94 & 1.53\\
Hybrid & UperNet SwinT~\cite{CAI2025104429} & 84.73 & 59.941 & 811.59 & ~9.36 & 1.778 & 0.65 & 0.10\\
Hybrid & UperNet SwinL~\cite{CAI2025104429} & \underline{85.63} & 233.962 & 825.92 & \underline{10.26} & 2.369 & 0.52 & 0.08\\
Transformer & SegFormer MiT-B0~\cite{CAI2025104429} & 75.37 & ~~~\textbf{3.752} & \textbf{20.89} & Baseline & \textbf{0.574} & \text{Baseline} & \textbf{13.62}\\
\bottomrule
\end{tabular}
\vspace{-.1cm}
\begin{flushleft} \footnotesize{Note: Input image size - $1000 \times 1000$. $\uparrow$ – higher is better; $\downarrow$ – lower is better; Boldface indicates the best result; underline indicates the 2nd-best.}\end{flushleft}
\end{center} \vspace{-0.2cm}
\end{table*}
\begin{table*}[!tp]
\setlength{\tabcolsep}{5pt} 
\caption{Model Efficiency on the Phenobench dataset}
\label{table-Phenobench-Efficiency} \vspace{-0.4cm}
\begin{center}
\begin{tabular}{llS[table-format=2.2]SS[table-format=3.3]cccS[table-format=2.2]}
\toprule 
\multicolumn{1}{c}{\textbf{Type}} & \multicolumn{1}{c}{\textbf{Model}} & \textbf{mIoU ($\%$) $\uparrow$} & \multicolumn{1}{c}{\textbf{Param. (M) $\downarrow$}} & \multicolumn{1}{c}{\textbf{GFLOPs} $\downarrow$} & \textbf{Diff$_{miou}$ $\uparrow$} & l\textbf{og(Param.) $\downarrow$} & \multicolumn{1}{c}{\textbf{Efficiency $(\%)$ $\uparrow$}} & \multicolumn{1}{c}{\textbf{FPS $\uparrow$}}\\
\midrule
\rowcolor{green!15} 
CNN & Our model (DAS-SK) & \textbf{85.55} & ~~\textbf{10.678} & \textbf{45.00} & \textbf{4.67} & \textbf{1.028} & \textbf{10.09} & 10.33\\
CNN & UNet ResNet34 \cite{ronneberger2015u} & 85.48 & 24.436 & 125.63 & 4.60 & 1.388 & \underline{2.64} & \underline{15.19}\\
CNN & DeepLabV3+ ResNet101 \cite{chen2017rethinking} & \underline{85.52} & 45.670 & 233.89 & \underline{4.64} & 1.660 & 1.25 & 7.75\\
CNN & DeepLabV3 ResNet101 \cite{chen2017rethinking} & 84.98 & 58.626 & 964.56 & 4.10 & 1.768 & 0.24 & 2.42\\
CNN & PSPNet ResNet50 \cite{zhao2017pyramid} & 80.88 & ~~\underline{24.314} & \underline{46.87} & Baseline & \underline{1.386} & \text{Baseline} & \textbf{22.05}\\
\bottomrule
\end{tabular}
\vspace{-.1cm}
\begin{flushleft} \footnotesize{Note:  Input image size - $1024 \times 1024$. $\uparrow$ – higher is better; $\downarrow$ – lower is better; Boldface indicates the best result; underline indicates the 2nd-best.}\end{flushleft}
\end{center} \vspace{-0.2cm}
\end{table*}


\noindent
A comprehensive summary of class-wise IoU scores and model performance bubble charts for all three datasets can be found in the Supplementary Materials. Table~\ref{table-Landcover-Efficiency}-\ref{table-Phenobench-Efficiency} results demonstrate that DAS-SK achieves an optimal balance between segmentation accuracy, computational efficiency, and inference speed. On Landcover.ai, DAS-SK attains 86.25\% mIoU, second only to the ensemble UNet, while requiring only 10.7M parameters and 11.25 GFLOPs, resulting in the highest efficiency and excellent real-time FPS. Similarly, on VDD, DAS-SK achieves 79.45\% mIoU with the highest efficiency and second-best FPS. On PhenoBench, DAS-SK reaches the best mIoU while remaining the most compact and efficient model, demonstrating strong generalization to fine-grained crop/weed segmentation.

Model selection under practical constraints further highlights the advantage of DAS-SK. In memory-limited scenarios (about 10–12M parameters), DAS-SK provides the best trade-off between accuracy and resource usage, outperforming lightweight alternatives such as BiSeNet or SegFormer MiT-B0, which either compromise accuracy or achieve only marginal gains in FPS. Larger transformers, hybrid or ensembles, while slightly improving mIoU, require $\geq$30M parameters and achieve \textless5 FPS, making them unsuitable for real-time deployment on UAVs or edge devices.


\subsection{Qualitative analysis}
\noindent
\begin{figure}[!t]
\centering  
\includegraphics[trim={0.00cm, 0.0cm, 0.00cm, 0.0cm}, clip, width=\columnwidth]{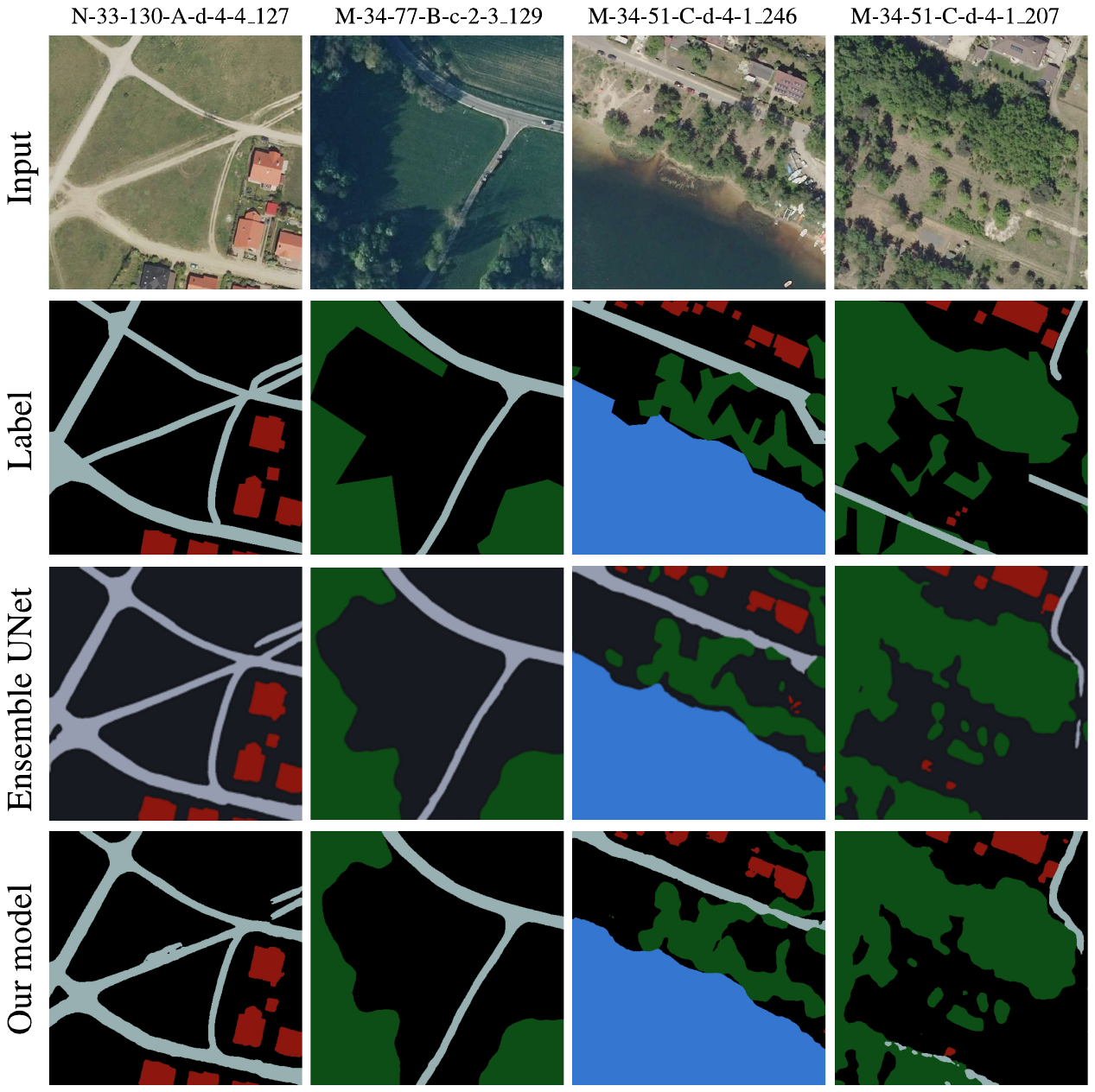} \vspace{-0.4cm}
\caption{Four prediction samples on LandCover.ai's test set.}
\label{fig-viz_landcover_ensembleUnet} \vspace{-0.2cm}
\end{figure}
\begin{figure}[!t]
\centering  
\includegraphics[trim={0.00cm, 0.0cm, 0.00cm, 0.0cm}, clip, width=\columnwidth]{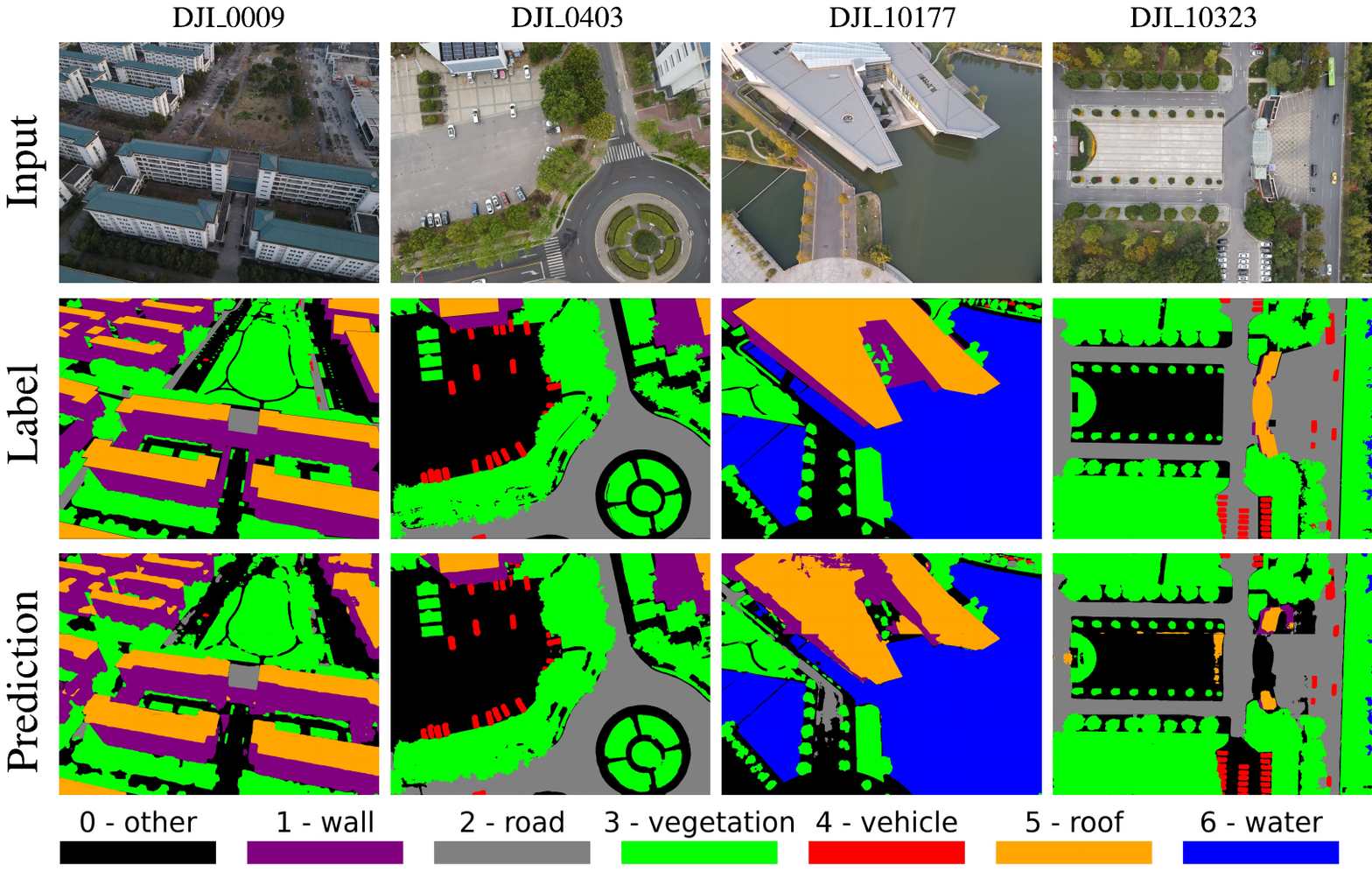} \vspace{-0.4cm}
\caption{Four predication samples on VDD's test set.}
\label{fig-viz_VDD}\vspace{-0.2cm}
\end{figure}

\begin{figure}[!t]
\centering  
\includegraphics[trim={0.00cm, 0.0cm, 0.00cm, 0.0cm}, clip, width=\columnwidth]{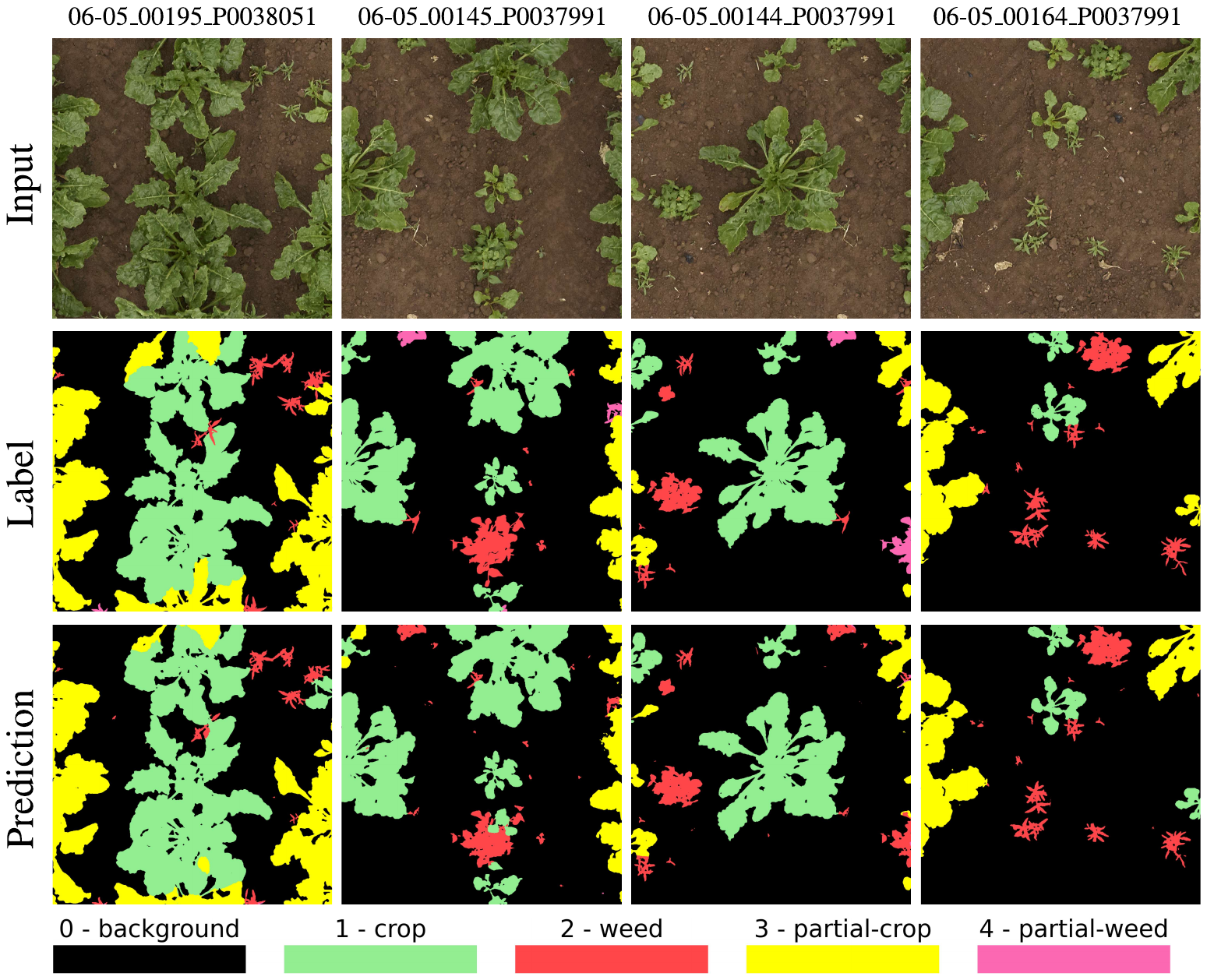}
\vspace{-0.4cm}
\caption{Four predication samples on PhenoBench's validation set.}
\label{fig-viz_pheno} \vspace{-0.2cm}
\end{figure}

\noindent
Across the three datasets (See Fig.~\ref{fig-viz_landcover_ensembleUnet}~to~\ref{fig-viz_pheno}), the proposed DAS-SK model consistently delivers accurate and visually reliable segmentation results. On LandCover.ai, it produces sharper building boundaries and smoother vegetation compared to Ensemble UNet. An additional quantitative visualization is provided in the Supplementary Materials. On the VDD dataset, DAS-SK demonstrates strong alignment with ground-truth labels, effectively delineating complex urban scenes and maintaining robustness under varying perspectives and lighting conditions. Similarly, on PhenoBench, the model accurately distinguishes crops, weeds, and partial vegetation classes even in cluttered or occluded environments, showing high precision in separating visually similar regions. Overall, DAS-SK achieves consistent, detailed, and context-aware segmentation across diverse scenes, confirming its generalization capability.

\section{Conclusion}
\label{Sec-Conclusion}
\noindent
In conclusion, this study presented DAS-SK, a lightweight and efficient semantic segmentation framework for high-resolution agricultural imagery. By integrating DAS and SK convolutions within a modified DeepLabV3 architecture featuring context-aware attention and multi-scale fusion, DAS-SK achieves an optimal precision-efficiency balance. Evaluations on LandCover.ai, VDD, and PhenoBench datasets show that DAS-SK consistently outperforms CNN-, transformer-, and hybrid-based baselines, demonstrating strong generalization and scalability while preserving a compact design, offering a practical and high-performing solution for agricultural and remote sensing applications. Future work will extend DAS-SK toward self-supervised and domain-adaptive segmentation under limited labels of agricultural imagery.

\section{Appendix}


This section provides additional technical details, implementation insights, and analyses that complement the main manuscript. Section~\ref{Sec-ablation-studies-appendix} presents the ablation studies, including training and validation performance analyses to evaluate convergence stability and the model’s overall learning behavior. Section~\ref{Sec-athematical-expressions} details the mathematical formulations of the proposed \texttt{DAS-SKConv} module, the enhanced Atrous Spatial Pyramid Pooling (ASPP) module, and the modified decoder structure. Section~\ref{sec-quantitative-appendix} provides a comprehensive quantitative evaluation across multiple datasets, while Section~\ref{sec-add-qualitative} offers an additional qualitative analysis to visually assess segmentation consistency and robustness. Together, these sections aim to provide an understanding of the model’s architectural design, ablation decisions, and performance of the proposed framework across the benchmark datasets.

\subsection{Additional ablation studies}
\label{Sec-ablation-studies-appendix}

\begin{table*}[!tp]
\setlength{\tabcolsep}{1pt} 
\caption{Summary of Ablation Studies} \vspace{-0.35cm}
\label{table-ablation-studies} 
\begin{center}
\begin{tabular}{c L{0.55\linewidth} S[table-format=2.2]SS[table-format=3.3]S[table-format=4.2]}
\toprule 
\multicolumn{1}{c}{\textbf{Stage}} & \multicolumn{1}{c}{\textbf{Description}} & \textbf{mIoU ($\%$)} & \multicolumn{1}{c}{\textbf{Param. (M)}} & \multicolumn{1}{c}{\textbf{GFLOPs}} & \multicolumn{1}{c}{\textbf{Memory (MB)}} \\
\midrule
1 & Baseline model (MobileNetV3-Large DeepLabV3) & 82.77 & 11.025 & 9.84 & 514.77 \\
2 & Replace all standard convolutions with dilation in the ASPP module with 
\texttt{DAS-Conv} of dilation rates of 4, 8, 12 and 24 and add skip connection (b7) from backbone. & 83.08 & 7.589 & 6.31 & 533.34 \\
3 & Add dilation rates of 18 and 26 and replace dilation rate of 24 with 22 & 83.23 & 7.606 & 6.33 & 533.34 \\
4 & Replace Global Average Pooling with Strip Pooling in ASPP & 83.34 & 7.852 & 6.83 & 554.31 \\
5 & Add a skip connection from base backbone to decoder (b4) & 83.58 & 7.898
& 7.03 & 574.39 \\
6 & Add auxiliary backbone (EfficientNet-B3) & 84.57 & 10.077 & 10.16	& 1549.56 \\
7 & Add a skip connection from auxiliary backbone (a4) & 84.90 & 10.084 & 10.19 &	1551.17 \\
8 & Replace \texttt{DAS-Conv} with \texttt{DAS-SKConv} in ASPP & 85.18 & 10.383 & 10.42 & 1577.54 \\
9 & Add a skip connection from base backbone (b2) & 85.23 & 10.411 & 10.90 & 1640.96 \\
10 & Add a skip connection from auxiliary backbone (clone(a3)) & 85.30 & 10.415 & 10.98 & 1645.17 \\
11 & Replace first separable conv right after Enhanced ASPP with a parallel conv & 85.79 & 10.678 & 11.25 & 1646.22 \\
12 & Add additional data augmentation of ShiftScaleRotate & 86.25 & 10.678 & 11.25 & 1646.22 \\
\bottomrule
\end{tabular}
\end{center}
\end{table*}


\begin{figure*}[!t]
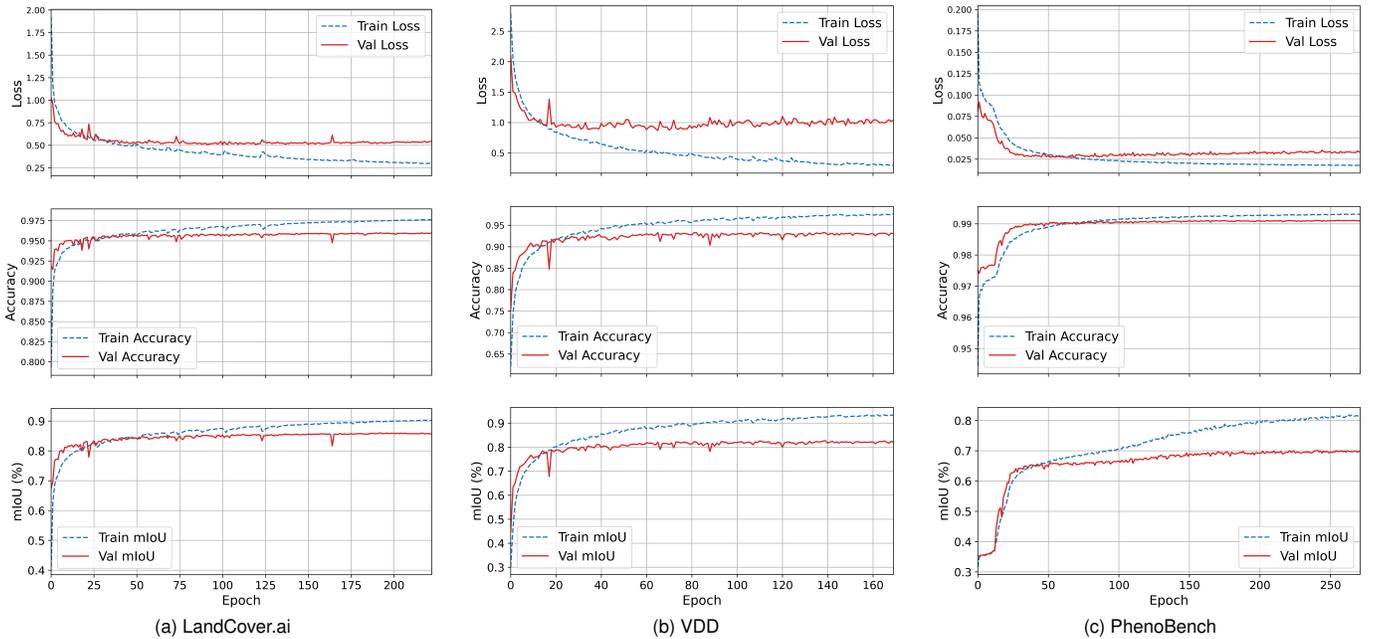

    \centering
    \subfloat[LandCover.ai]{%
        \includegraphics[height=8cm]{Figures/val_curves_landcover.png}%
        \label{fig:subfig1}
    }
    \hfill
    \subfloat[VDD]{%
        \includegraphics[height=8cm]{Figures/val_curves_VDD.png}%
        \label{fig:subfig2}
    }
    \hfill
    \subfloat[PhenoBench]{%
        \includegraphics[height=8cm]{Figures/val_curves_pheno.png}%
        \label{fig:subfig3}
    }
    \caption{Training and validation performance curves for LandCover.ai, VDD, and PhenoBench datasets.}
    \label{fig-performance-curves-appendix}
    \vspace{-0.2cm}
\end{figure*}

As PhenoBench requires submitting model predictions for its test set to obtain results, the LandCover.ai dataset was instead used as a trial platform to improve the DAS model. Numerous experiments were conducted to modify the model architecture and enhance performance; however, only those that led to significant improvements are detailed in Table~\ref{table-ablation-studies}.

Starting from the baseline MobileNetV3-Large DeepLabV3, which achieves an mIoU of 82.77\% with 11.0M parameters and 9.84 GFLOPs, successive modifications demonstrate a careful balance between accuracy gains and resource costs.

Replacing standard convolutions with \texttt{DAS-Conv} and adding a skip connection from backbone (b7) in Stage 2 significantly reduced computational cost—parameters dropped by 31\% and GFLOPs by 36\% while improving mIoU by +0.31\%. The integration of multi-dilation receptive fields through DAS-Conv enhanced contextual understanding with lower computation, while the added skip connection from the backbone (b7) further strengthened high-level semantic guidance to the ASPP, contributing to improved feature consistency and better boundary delineation. Together, these changes delivered a highly favorable impact-to-cost ratio, showing that structural efficiency can coexist with accuracy improvements.

Stage 3 involved adjustment of the dilation configuration, which yielded an additional +0.15\% mIoU gain with negligible cost increase, suggesting optimal receptive field tuning. Substituting global average pooling with strip pooling in Stage 4 improved spatial context aggregation, resulting in an mIoU gain of +0.11\% at a small cost increase of 0.25M parameters and 0.5 GFLOPs. In Stage 5, adding backbone skip connections from deeper layers further enhanced feature fusion, achieving an mIoU gain of +0.24\% with minimal overhead. To accommodate these additional inputs from the backbones to the decoder at this and subsequent stages in the ablation studies, channel-wise concatenation followed by a separable convolution, ReLU activation, batch normalization, and bilinear upsampling was implemented.

Incorporating an auxiliary EfficientNet-B3 backbone in Stage 6 led to a substantial mIoU gain of +0.99\%, albeit with increases in parameters (+27\%) and memory usage (about 3× higher). Despite the higher cost, this step provided one of the most notable gains, highlighting the auxiliary encoder’s complementary feature representation. The subsequent addition of a skip connection (a4) from this auxiliary path in Stage 7 added marginal cost but yielded a further +0.33\% mIoU gain, confirming the benefit of cross-scale fusion.

Replacing DAS-Conv with DAS-SKConv in Stage 8 brought another meaningful +0.28\% mIoU gain with moderate computational growth (+3\% parameters, +2\% GFLOPs), underscoring the SK module’s adaptivity to multi-scale feature variation. Additional skip connections (b2 and clone(a3)) from both main and auxiliary backbones in Stages 9-10 provided smaller incremental improvements of +0.05\% and +0.07\% mIoU gains, respectively, at slight cost increases, suggesting diminishing returns at this point. Initial experiment with a direct skip connection from $a3$ to the decoder resulted in lower performance due to backpropagation affecting the weights of the auxiliary backbone via $a3$. To prevent unintended weight updates in the auxiliary backbone, a copy of the $a3$ feature maps was used instead, ensuring that the auxiliary backbone from $a3$ backwards remained unchanged during training.

In Stage 11, replacing the post-ASPP separable convolution with a parallel convolution yielded a larger +0.49\% mIoU gain for a modest 2.5\% increase in computational cost, indicating an efficient trade-off. Finally, the application of ShiftScaleRotate data augmentation in Stage 12 enhanced the final performance with an mIoU gain of +0.46\% without adding computational or parameter burden, emphasizing the crucial role of data diversity in generalization.

In summary, the progression illustrates that early architectural redesigns (Stages 2–5) achieved high efficiency gains with minimal resource increase, while later stages (6–11) focused on refined multi-scale and multi-backbone feature fusion to maximize accuracy. The final configuration balances strong performance with acceptable efficiency, demonstrating the effectiveness of the DAS-SK architecture in optimizing the impact-to-cost ratio across design stages.

It is worth noting that the above architectural refinements are based on a $512\times512$ image input. For larger input sizes, such as those in the VDD and PhenoBench datasets, the number of \texttt{DAS-SKConv} and their dilation rates can be further fine-tuned to achieve broader receptive fields and potentially better performance—an aspect that will be explored in future work. For simplicity, the model refined on the LandCover.ai dataset is utilized for both VDD and PhenoBench in this study.

\subsubsection{Training and Validation Performance Curves}

To assess the training stability and generalization behavior of the proposed model, Fig.~\ref{fig-performance-curves-appendix} presents the training and validation curves for loss, accuracy, and mean Intersection-over-Union (mIoU) across the three evaluated datasets. These curves provide insight into the convergence dynamics and the model’s ability to maintain consistent performance between training and validation phases.

For LandCover.ai, early stopping was triggered at epoch 222, with the best validation mIoU of 0.8602 achieved at epoch 192.
For VDD, early stopping occurred at epoch 169, with the best validation mIoU of 0.8269 recorded at epoch 139.
For PhenoBench, early stopping was applied at epoch 271, with the best validation mIoU of 0.7016 attained at epoch 241.

Across all datasets, the training losses exhibit smooth convergence, while validation losses stabilize early, indicating effective and consistent learning. Accuracy and mIoU curves demonstrate steady improvement and close alignment between training and validation phases, reflecting stable optimization behavior and minimal overfitting throughout training.

\subsection{Mathematical expressions}
\label{Sec-athematical-expressions}

\subsubsection{DAS-SKConv module}

\begin{figure*}[!t]
\centering  
\includegraphics[trim={0.00cm, 21.60cm, 0.00cm, 0.0cm}, clip, width=1.0\textwidth]{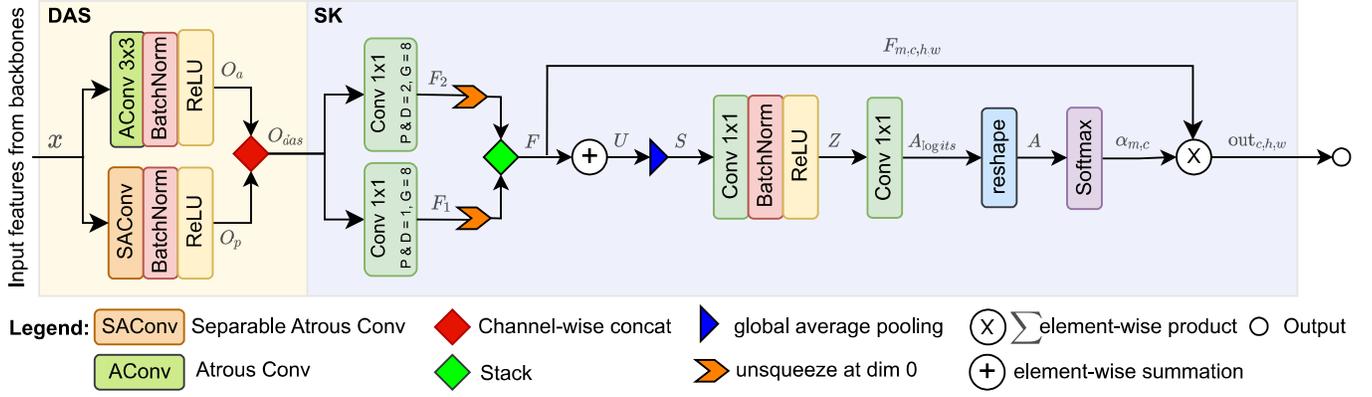} \vspace{-0.4cm}
\caption{The proposed \texttt{DAS-SKConv} module. The DAS block combines atrous separable and standard atrous convolutions to capture both fine and broad spatial features. The SK attention mechanism adaptively weights multi-branch features through channel-wise attention, producing context-aware representations.}
\label{fig-DAS-SKConv-appendix} \vspace{-0.2cm}
\end{figure*}

\noindent
The \texttt{DAS-SKConv} module in Fig.\ref{fig-DAS-SKConv-appendix} is designed to enhance feature extraction by combining multiscale spatial context with adaptive channel attention. It begins with a DAS block that processes the input feature maps through two parallel convolution paths, which are as follows: 

In the atrous separable branch, the input $x \in \mathbb{R}^{C_{in} \times H \times W}$ first undergoes a depthwise atrous convolution:
\begin{equation}
O_{da} = \text{Conv}^{dw}_{3,d}(x) \in \mathbb{R}^{ C_{\text{in}}\times H\times W},
\end{equation}
where each input channel is convolved independently with a $3 \times 3$ kernel, dilation $d$, and padding $d$. This is followed by a pointwise convolution for channel mixing and reduction:
\begin{equation}
O_{p}  = \text{Conv}_{1\times1}(O_{da}) \in \mathbb{R}^{[C_{\text{out}}/2] \times H\times W},
\end{equation}
and then batch normalization with ReLU activation: 
\begin{equation}
O_{p}  = \text{ReLU}(\text{BN}(O_{p})).
\end{equation}

In the standard atrous branch, a regular dilated convolution is applied:
\begin{equation}
O_{a} = \text{Conv}_{3,d}(x) \in \mathbb{R}^{[ C_{\text{out}}/2] \times H\times W},
\end{equation}
with kernel size $3 \times 3$, dilation $d$, and padding $d$, followed again by batch normalization and ReLU:
\begin{equation}
O_{a} = \text{ReLU}(\text{BN}(O_{a})).
\end{equation}

The final output of the DAS block is obtained by concatenating the two feature maps along the channel dimension:
\begin{equation}
O_{\text{das}} = \text{Concat}(O_{p}, O_{a}) \in \mathbb{R}^{C_{\text{out}}\times H\times W}.
\end{equation}

These concatenated features are then passed through an SK attention mechanism, which adaptively recalibrates the channel-wise feature responses. The SK attention module learns to selectively emphasize or suppress features from different receptive fields, allowing the network to dynamically focus on the most informative spatial scales for each input.
Each SK branch convolution produces
\begin{equation}
F_m = \text{Conv}_m(O_{\text{das}}) \in \mathbb{R}^{C\times H\times W}, \quad m=1,\dots,M.
\end{equation}
which are stacked as
\begin{equation}
\mathcal{F} = \text{stack}(F_1,\dots,F_M) \in \mathbb{R}^{M\times C\times H\times W}.
\end{equation}
The fused representation is obtained by summation,
\begin{equation}
U = \sum_{m=1}^{M} F_m \in \mathbb{R}^{C\times H\times W}.
\end{equation}
A global average pooling is applied,
\begin{equation}
S = \text{GAP}(U) \in \mathbb{R}^{C\times 1 \times 1},
\end{equation}
followed by channel reduction and nonlinearity,
\begin{equation}
Z = \text{ReLU}\!\left(\text{BN}\!\left(\text{Conv}_{1\times1}(S)\right)\right) \in \mathbb{R}^{d\times 1\times 1},
\end{equation}
\begin{equation}
\quad d = \max(C/r, L).
\end{equation}
Branch-channel attention logits are generated as
\begin{equation}
A_{\text{logits}} = \text{Conv}_{1\times1}(Z) \in \mathbb{R}^{M\cdot C\times 1\times 1},
\end{equation}
reshaped to $A \in \mathbb{R}^{M \times C \times 1 \times 1}$ and normalized with a softmax across the branch dimension:
\begin{equation}
\alpha_{m,c} = \frac{\exp(A_{m,c})}{\sum_{m'=1}^{M}\exp(A_{m',c})}.
\end{equation}
The final output is then the attention-weighted sum of branch features,
\begin{equation}
\text{out}_{c,h,w} = \sum_{m=1}^{M} \alpha_{m,c}\cdot F_{m,c,h,w},
\end{equation}
yielding an output features $\in \mathbb{R}^{C\times H\times W}$.

This combination of dual atrous convolutional sampling and adaptive selective-kernel attention enables \texttt{DAS-SKConv} to provide rich, context-aware feature representations, making it effective for semantic segmentation and other dense prediction tasks where both local detail and global context are crucial.

\subsection {Enhanced ASPP module}
\label{subsection-assp-appendix}

\begin{figure}[!tp]
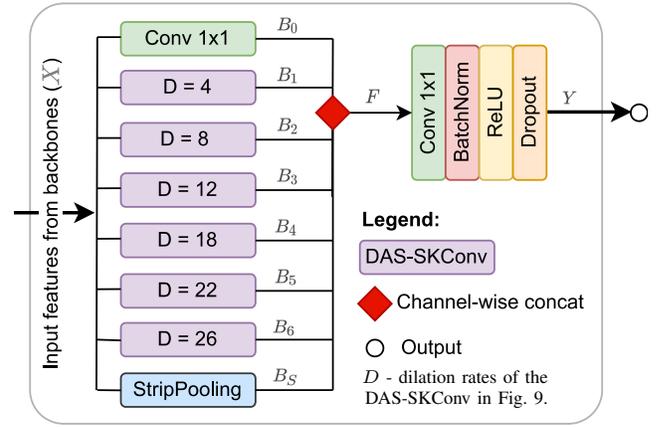

\centering
\begin{overpic}[trim={0.5cm 13.7cm 0.0cm 0.0cm}, clip, width=0.95\columnwidth]{Figures/enhanced_aspp.pdf}
  \put(55,4){\scriptsize \shortstack[r]{
$D$ - dilation rates of the \\ DAS-SKConv
in Fig.~\ref{fig-DAS-SKConv-appendix}.}}
\end{overpic} \vspace{-0.2cm}
\caption{The enhanced ASPP module. High-dimensional backbone features are processed via parallel branches, including a $1\times1$ Conv, six \texttt{DAS-SKConv} with varying dilation rates, and a strip pooling branch. 
}
\label{fig-enhanced-ASPP-module-appendix} \vspace{-0.2cm}
\end{figure}

The Enhanced ASPP module, as shown in Fig.\ref{fig-enhanced-ASPP-module-appendix}, is designed to enrich feature representations by capturing contextual information at multiple scales. It begins by taking a high-dimensional feature map from both primary and auxiliary backbone networks, denoted as $X \in \mathbb{R}^{960 \times H \times W}$. To capture diverse contextual information, this module processes $X$ through multiple parallel branches.

First, a $1\times1$ convolution reduces the channel dimension:
\begin{equation}
B_o=\text{Conv}_{1\times1}(X) \in \mathbb{R}^{256 \times H\times W}.
\end{equation}

In parallel, six DAS\_SK convolutions with kernel size $3\times3$ and dilation rates $d \in \{4,8,12,18,22,26\}$ are applied:
\begin{equation}
B_i = \text{DAS\_SKConv}_{3 \times 3, d_i}(X),
\end{equation}
where $B_i \in \mathbb{R}^{128\times H \times W}$ for $i=1,\dots,6$.

To further improve the module’s ability to model long-range dependencies, a strip pooling branch is included. Unlike standard pooling, strip pooling captures context along horizontal and vertical strips, which is efficient for structured and elongated objects such as vegetation rows.
\begin{equation}
B_s = \text{StripPool}(X) \in \mathbb{R}^{256 \times H \times W}.
\end{equation}

All branch outputs are concatenated along the channel dimension, resulting in a fused feature map:
\begin{equation}
F = \text{Concat}(B_0,B_1, \dots, B_6, B_s) \in \mathbb{R}^{1280 \times H \times W}.
\end{equation}

Finally, to make this representation compact and more efficient for subsequent processing, a $1\times1$ convolution is applied to reduce the dimensionality back to 256 channels. This is followed by batch normalization, ReLU activation, and dropout to stabilize training and prevent overfitting.
\begin{equation}
Y = \text{Dropout}(\text{ReLU}(\text{BN}(\text{Conv}_{1 \times 1}(F))))
\end{equation}
yielding an output features $\in \mathbb{R}^{256\times H\times W}$.

Overall, the Enhanced ASPP module improves the representational power of the backbone features by combining local details, multi-scale context, and global structural information. The resulting output feature map is compact and context-rich, making it highly efficient for downstream segmentation tasks.

\subsection{Decoder}

\begin{figure*}[!ht]
\centering   
\includegraphics[trim={0.00cm, 23.8cm, 0.00cm, 0.0cm}, clip, width=1.0\textwidth]{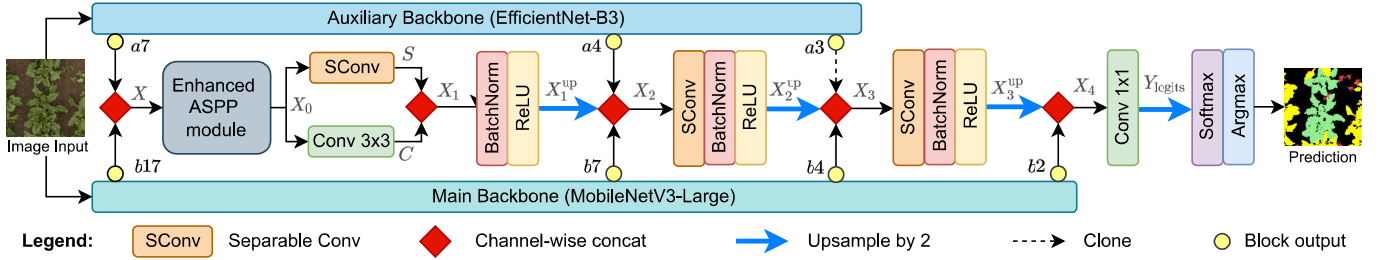} \vspace{-0.4cm}
\caption{The overall architecture. The backbones extract multi-scale features from the input, which are then fused and refined via the enhanced ASPP with \texttt{DAS-SKConv} (cf.~Fig.~\ref{fig-DAS-SKConv-appendix}). The decoder then progressively upsamples and refines the fused features with skip connections to produce accurate segmentation.}
\label{fig-overall-architecture-appendix} \vspace{-0.2cm}
\end{figure*}

The decoder, illustrated in Fig.~\ref{fig-overall-architecture-appendix}, begins by concatenating the high-level feature maps $a7$ and $b17$ from the backbones channel-wise to form $X \in \mathbb{R}^{960 \times H \times  W}$. It is then enriched with contextual information using the Enhanced ASPP module, resulting in $X_{0} \in \mathbb{R}^{256 \times H \times W}$. 

To further refine these local and multi-scale contextual features, $X_{0}$ is passed through two parallel operations: a separable convolution (SConv) and a standard $3\times3$ convolution.
\begin{equation}
S = \text{SConv}(X_0) \in \mathbb{R}^{128 \times H \times W},
\end{equation}
\begin{equation}
C = \text{Conv}_{3 \times 3}(X_0) \in \mathbb{R}^{128 \times H \times W}.
\end{equation}

The results are concatenated along the channel dimension, producing a strong combined representation:
\begin{equation}
X_1 = \text{Concat}(S,C) \in \mathbb{R}^{256 \times H \times W}
\end{equation}
Next, the learned features undergo batch normalization and ReLU activation. 
\begin{equation}
X_1' = \text{ReLU}(\text{BN}(X_1))
\end{equation}

Next, the decoder progressively upsamples the features to recover spatial detail and aligns them with lower-level features from the backbone and auxiliary backbone. The first upsampling step doubles the spatial resolution:
\begin{equation}
X_1^{up}=\text{Upsample}(X_1', scale=2) \in \mathbb{R}^{256 \times 2H \times 2W}.
\end{equation}

This upsampled feature map is concatenated with the corresponding feature map from the backbones ($a4$ and $b7$), forming a richer representation:
\begin{equation}
X_2 = \text{Concat}(X_1^{up}, a4, b7) \in \mathbb{R}^{344 \times 2H \times 2W}.
\end{equation}

A separable convolution followed by batch normalization and ReLU activation, denoted as $\phi(.)$, then refines this merged feature set:
\begin{equation}
X_2' = \phi(X_2) \in \mathbb{R}^{148 \times 2H \times 2W}.
\end{equation}

The resulting feature map is upsampled, fused with backbone outputs of $a3$ and $b4$, and refined again:
\begin{equation}
X_2^{up} = \text{Upsample}(X_2', scale =2) \in \mathbb{}^{148 \times 4H \times 4W},
\end{equation}
\begin{equation}
X_3=\text{Concat}(X_2^{up}, \text{clone}(a3), b4) \in \mathbb{R}^{204 \times 4H \times 4W},
\end{equation}
\begin{equation}
X_3' = \phi(X_3) \in \mathbb{R}^{148 \times 4H \times 4W}.
\end{equation}

Finally, the decoder upsamples the output feature representation once more, concatenates with the shallowest primary backbone output ($b2$), and refines it using separable convolution, batch normalization, and ReLU.
\begin{equation}
X_3^{up} = \text{Upsample}(X_3', scale =2) \in \mathbb{}^{148 \times 8H \times 8W},
\end{equation}
\begin{equation}
X_4=\text{Concat}(X_3^{up}, b2) \in \mathbb{R}^{164 \times 8H \times 8W},
\end{equation}
\begin{equation}
X_4' = \phi(X_4) \in \mathbb{R}^{164 \times 8H \times 8W}.
\end{equation}

A final $1\times1$ convolution projects the features into $C$ channels, which corresponds to the number of target classes, and it is then resized into class logits:
\begin{equation}
Y=\text{Conv}_{1 \times 1}(X_4') \in \mathbb{R}^{C \times 8H \times 8W},
\end{equation}
\begin{equation}
Y_{logits}=\text{Upsample}(Y, \text{scale}=2) \in \mathbb{R}^{C \times 16H \times 16W}.
\end{equation}

The segmentation map can be obtained by applying the Softmax activation function and argmax to $Y_{logits}$. 

\subsection{Quantitative analysis}
\label{sec-quantitative-appendix}

\begin{table*}[ht!]
\centering
\begin{minipage}[t]{0.5\textwidth} 
    \centering
    \caption{Performance Analysis on LandCover's Test Set}
    \label{table-landcover-miou} \vspace{-0.2cm}
    \setlength{\tabcolsep}{3pt} 
    \begin{tabular}{lc|ccccc}
    \toprule 
    \multicolumn{1}{c}{\multirow{2}*{\textbf{Model}}} & \textbf{mIoU $ \uparrow$} & \multicolumn{5}{c}{\textbf{Class-wise IoU ($\%$) $ \uparrow$}}\\ 
    & ($\%$) & BG & Build & WoodLnd. & Water & Rd.\\
    \midrule
    Ensemble UNet & \textbf{88.02} & \textbf{94.30} & \textbf{85.33} & \textbf{92.43} & \textbf{95.42} & \textbf{72.62} \\
    \rowcolor{green!15} 
    Our model (DAS-SK)  & \underline{86.25} & \underline{93.42} & \underline{82.16} & 91.29 & \underline{94.72} & \underline{69.68} \\
    MA-DBFAN & 85.30 & \textbf{94.30} & 79.20 & \underline{91.30} & 94.70 & 67.20 \\
    SegFormer MiT-B2 & 84.40 & 93.20 & 76.90 & 91.20 & 94.40 & 66.30 \\ 
    Diff-HRNet & 84.22 & 93.00 & 79.10 & 90.10 & 93.10 & 65.80 \\
    UNet & 83.40 & 92.40 & 77.80 & 90.10 & 93.50 & 63.40 \\
    DeepLabV3 & 83.00 & 92.30 & 76.70 & 90.10 & 92.40 & 63.60 \\
    TransUNet & 82.90 & 92.10 & 77.90 & 89.90 & 92.50 & 62.40 \\
    HRNet & 82.80 & 92.10 & 76.50 & 89.90 & 92.70 & 62.60 \\
    BiSeNet & 80.50 & 91.00 & 71.80 & 88.60 & 91.20 & 59.60 \\
    \bottomrule
    \end{tabular}
    \vspace{0.15cm}
\begin{center} {\scriptsize 
$\uparrow$ - higher is better, $\downarrow$ - lower is better, Boldface - the best, underline - 2nd-best results. \\ BG- Background, Build- Buildings, WoodLnd.- Woodland, Rd.- Road, Veg.- Vegetation. }\end{center}

\end{minipage}%
\begin{minipage}[t]{0.5\textwidth} 
    \centering
    \setlength{\tabcolsep}{6pt} 
    \caption{Performance Analysis on Phenobench' Test Set}
    \label{table-phenobench-miou} \vspace{-0.2cm}
    \begin{tabular}{lc|ccc}
    \toprule 
    \multicolumn{1}{c}{\multirow{2}*{\textbf{Model}}} & \textbf{mIoU $\uparrow$} & \multicolumn{3}{c}{\textbf{Class-wise IoU ($\%$) $\uparrow$}}\\
     & ($\%$) & Soil & Crop & Weed \\
    \midrule
    \rowcolor{green!15} 
    Our model (DAS-SK) & \textbf{85.55} & \textbf{99.34} & \textbf{94.16} & \underline{63.15} \\
    DeepLabV3+ ResNet101 & \underline{85.52} & 99.29 & \underline{94.00} & \textbf{63.28} \\
    UNet ResNet34 & 85.48 & \underline{99.31} & 93.97 & 63.14 \\
    DeepLabV3 ResNet101 & 84.98 & 99.13 & 93.00 & 62.81 \\
   PSPNet ResNet50 & 80.88 & 99.04 & 91.81 & 51.80 \\
    \bottomrule
    \end{tabular}\\ \vspace{0.2cm}

\setlength{\tabcolsep}{1.5pt} 
\caption{Performance Analysis on VDD's Test Set}
\label{table-VDD-miou} 
\begin{center} \vspace{-0.45cm}
\begin{tabular}{lc|ccccccc}
\toprule 
\multicolumn{1}{c}{\multirow{2}*{\textbf{Model}}} & \textbf{mIoU $\uparrow$} & \multicolumn{7}{c}{\textbf{Class-wise IoU ($\%$) $\uparrow$}}\\
& ($\%$) & Other & Wall & Rd. & Veg. & Vehicle & Roof & Water\\
\midrule
Mask2Former & \textbf{83.21} & \textbf{75.76} & \underline{69.01} & \textbf{79.07} & \textbf{93.11} & \underline{74.25} & \textbf{94.27} & \textbf{97.02} \\
\rowcolor{green!15} 
Ours: DAS-SK & \underline{79.45} & \underline{66.51} & \textbf{71.77} & \underline{73.81} & \underline{90.09} & \textbf{74.94} & \underline{86.81} & \underline{92.22} \\
\bottomrule
\end{tabular}
\end{center}
\end{minipage}
\end{table*}

\begin{figure}[!t]
\centering  
\includegraphics[trim={0.00cm, 0.0cm, 0.00cm, 0.0cm}, clip, width=\columnwidth]{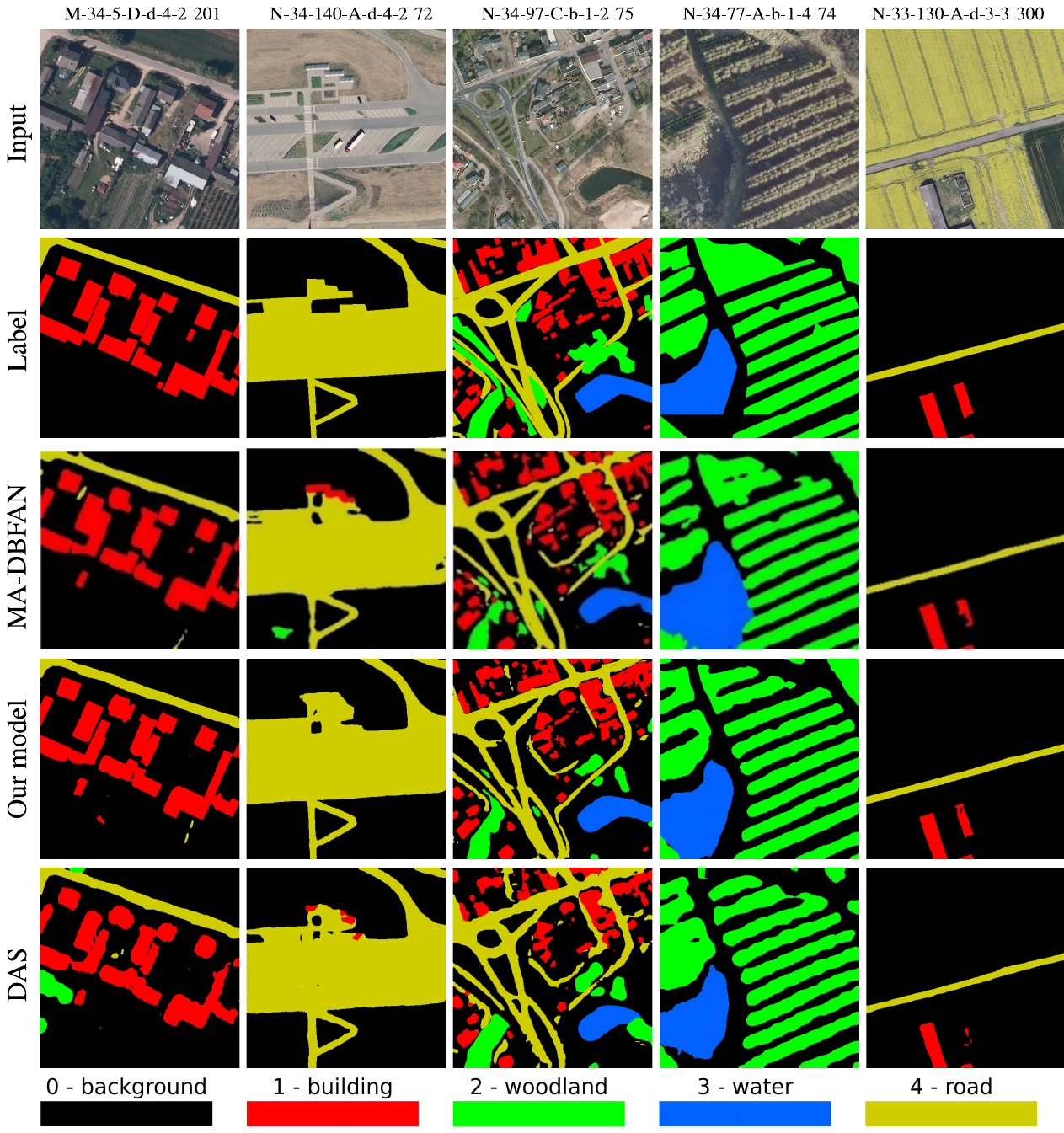}
\caption{Predication samples on LandCover.ai's test set.}
\label{fig-viz_landcover_MA-DBFAN}\vspace{-0.2cm}
\end{figure}

\noindent
Table \ref{table-landcover-miou}–\ref{table-VDD-miou} present the quantitative comparison of the proposed DAS-SK model against several state-of-the-art segmentation methods across three benchmark datasets: LandCover.ai, PhenoBench, and VDD. Overall, DAS-SK consistently demonstrates strong generalization and competitive accuracy while maintaining a lightweight and efficient architecture.

On the LandCover.ai dataset, DAS-SK achieves an overall mIoU of 86.25\%, ranking second only to the Ensemble-UNet, which relies on multi-model aggregation for performance gain. Despite this, DAS-SK surpasses most transformer- and CNN-based models such as SegFormer, Diff-HRNet, and DeepLabV3, highlighting the efficiency of the SK-enhanced convolutional design and refined decoder structure. The model performs particularly well in the Background, Building, and Water classes, indicating its robustness in delineating high-contrast and structurally diverse regions.

For the PhenoBench dataset, which contains fine-grained agricultural field imagery, DAS-SK attains the highest mIoU of 85.55\%, marginally outperforming DeepLabV3+and UNet ResNet34. The model achieves the best IoU scores for the Soil and Crop classes and a competitive score for Weed, demonstrating improved class discrimination and resilience against inter-class similarity. The superior performance here validates the model’s capacity to handle challenging agricultural scenes characterized by subtle texture and illumination variations.

On the VDD dataset, DAS-SK achieves an mIoU of 79.45\%, second only to Mask2Former. It excels in specific classes such as wall and vehicle, outperforming Mask2Former in these categories. Its consistently high IoU across heterogeneous classes—ranging from vegetative regions to man-made structures—underscores its adaptability to complex urban scenes, suggesting strong generalization beyond the training domain.

Across the three datasets, DAS-SK consistently demonstrates competitive performance, frequently achieving top two positions in mIoU and class-wise IoU metrics. Its performance profile suggests strong versatility: it handles both natural and urban scenes effectively while remaining computationally efficient. The results validate the effectiveness of integrating dual atrous separable convolutions with selective kernel mechanisms for capturing multi-scale contextual information and fine-grained details across diverse segmentation tasks.


\begin{figure*}[htbp]
    \centering
    \subfloat[LandCover.ai]{%
        \includegraphics[width=0.325\linewidth]{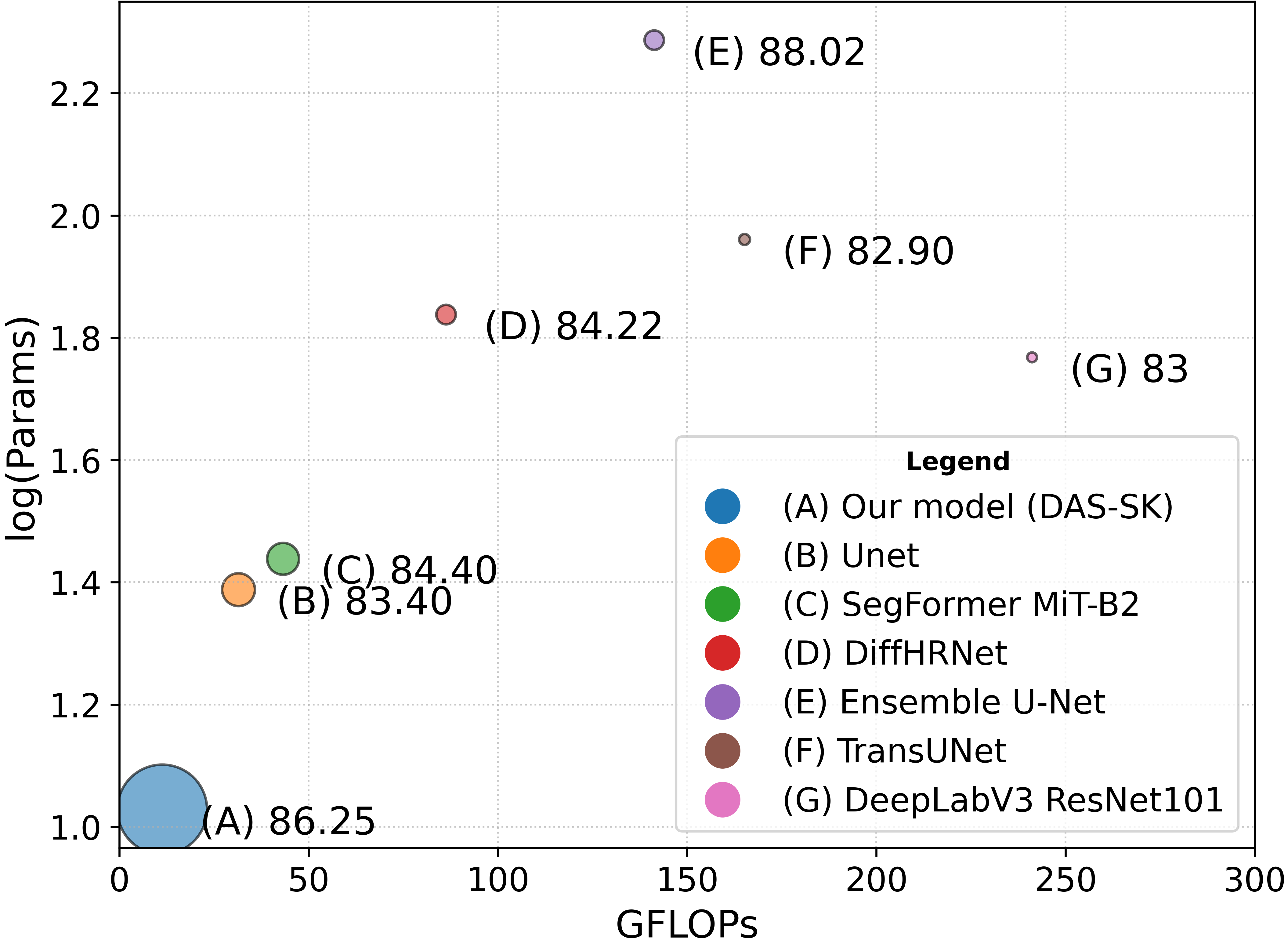}%
        \label{fig:subfig1p}
    }
    \subfloat[VDD]{%
        \includegraphics[width=0.325\linewidth]{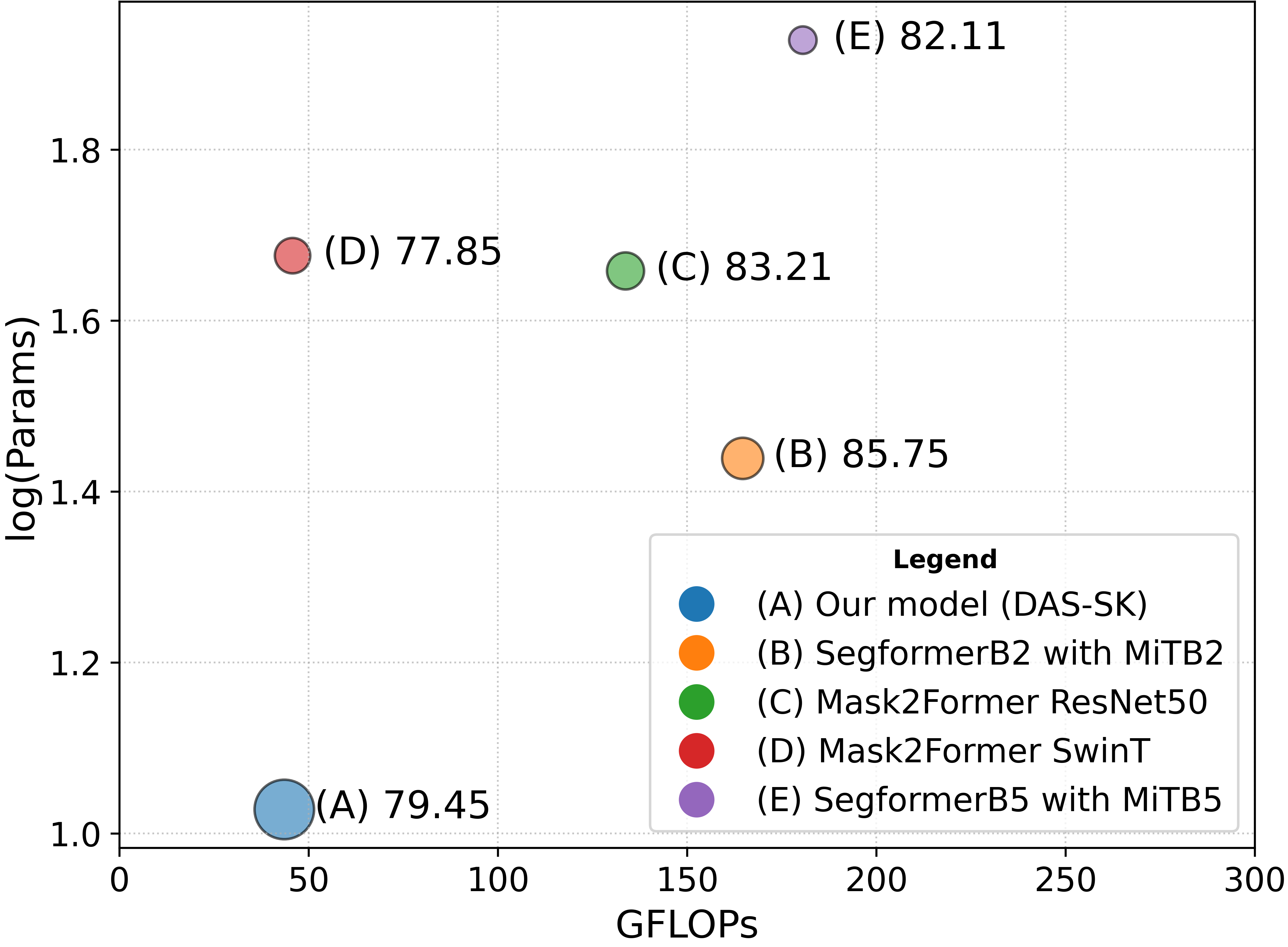}%
        \label{fig:subfig2p}
    }
    \subfloat[PhenoBench]{%
        \includegraphics[width=0.325\linewidth]{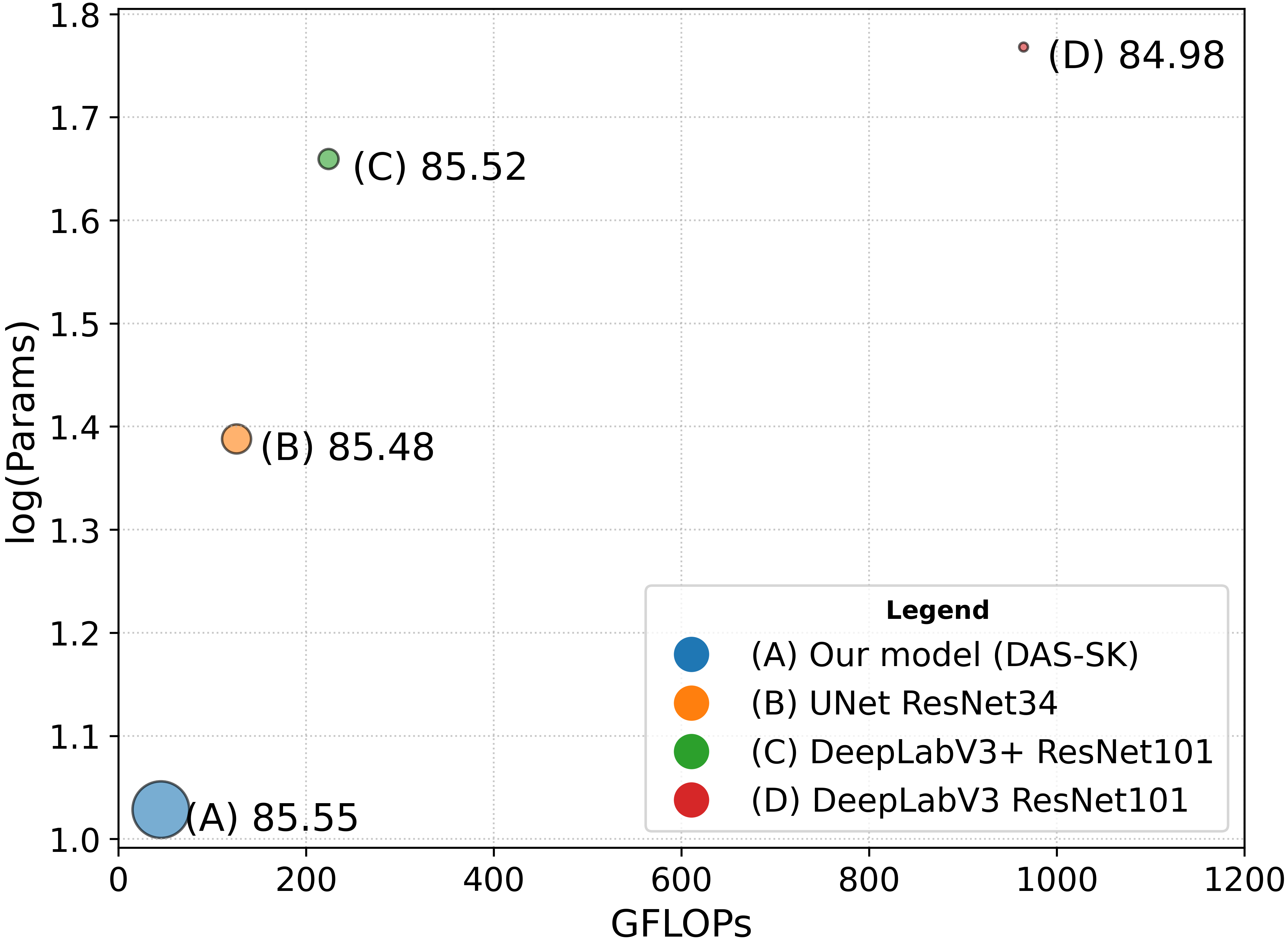}%
        \label{fig:subfig3p}
    }
    \caption{Bubble chart of model performance, with bubble size representing efficiency ($\%$) and values next to the bubble indicating mIoU ($\%$). Refer to Table VI-VIII in the main text; a few models are not included in these plots, either due to being a baseline or an outlier to the plot range.
    }
    \label{fig-subplots-efficiency} \vspace{-0.2cm}
\end{figure*}

Fig.~\ref{fig-subplots-efficiency} illustrates the efficiency of different segmentation models across the three datasets. The x-axis denotes GFLOPs and the y-axis the logarithm of parameters, with more lightweight models positioned toward the lower left. Bubble sizes reflect overall efficiency, and each is labeled with its mIoU ($\%$) to show the accuracy–efficiency trade-off. Across all subfigures, DAS-SK consistently appears near the lower-left corner while maintaining high mIoU, underscoring its strong balance between efficiency and segmentation quality—ideal for resource-constrained agricultural applications.

\subsubsection{Additional qualitative results}
\label{sec-add-qualitative}

Fig.~\ref{fig-viz_landcover_MA-DBFAN} presents a qualitative comparison between MA-DBFAN and our model across diverse remote sensing scenes in the LandCover.ai's test set. While MA-DBFAN captures the overall structure, it often struggles with fine boundaries and small objects. In contrast, our model demonstrates superior delineation of buildings, roads, and water bodies, while maintaining consistency in large homogeneous regions such as woodland and farmland. These results highlight the reliability of our approach in accurately segmenting both large-scale and detailed features in complex aerial imagery.

\bibliography{main} 
\bibliographystyle{IEEEtran} 

\end{document}